\documentclass[sigconf]{acmart}

\usepackage{soul}
\usepackage{balance}
\usepackage{pifont}
\usepackage{url}

\usepackage{graphicx}
\usepackage{amsmath} 
\usepackage{amsthm}
\usepackage{booktabs}
\usepackage{algorithm}
\usepackage{algorithmic}

\usepackage{bm}
\hypersetup{
	colorlinks=true,
	linkcolor=red,
	citecolor=blue,
} 
\usepackage{bbm}
\usepackage{colortbl} 
\usepackage{xcolor} 
\usepackage{array}   
\usepackage{arydshln}
\usepackage{subfigure}
\usepackage{amsmath}
\usepackage{natbib}  
\usepackage{caption} 

\renewcommand{\thefootnote}{\fnsymbol{footnote}}

\copyrightyear{2023} 
\acmYear{2023} 
\setcopyright{rightsretained} 
\acmConference[MM '23]{Proceedings of the 31st ACM International Conference
on Multimedia}{October 29-November 3, 2023}{Ottawa, ON, Canada}
\acmBooktitle{Proceedings of the 31st ACM International Conference on
Multimedia (MM '23), October 29-November 3, 2023, Ottawa, ON, Canada}
\acmDOI{10.1145/3581783.3611975}
\acmISBN{979-8-4007-0108-5/23/10}

\begin{document}

\title{SEAM: \underline{Se}arching Tr\underline{a}nsferable \underline{M}ixed-Precision Quantization Policy through Large Margin Regularization}

\author{Chen Tang}
\email{tc20@mails.tsinghua.edu.cn}
\affiliation{%
  \institution{SIGS \& Dept. of Computer Science and Technology, Tsinghua University}
}

\author{Kai Ouyang}
\email{oyk20@mails.tsinghua.edu.cn}
\affiliation{%
  \institution{SIGS, Tsinghua University}
}

\author{Zenghao Chai}
\email{zenghaochai@comp.nus.edu.sg}
\affiliation{%
  \institution{National University of Singapore}
}

\author{Yunpeng Bai}
\email{byp20@mails.tsinghua.edu.cn}
\affiliation{%
  \institution{SIGS, Tsinghua University}
}

\author{Yuan Meng$^{\dagger}$}
\email{yuanmeng@mail.tsinghua.edu.cn}
\affiliation{%
  \institution{Dept. of Computer Science and Technology, Tsinghua University}
}

\author{Zhi Wang$^{\dagger}$}
\email{wangzhi@sz.tsinghua.edu.cn}
\affiliation{%
  \institution{SIGS, Tsinghua University}
  \institution{Peng Cheng Laboratory}
}

\author{Wenwu Zhu$^{\dagger}$}
\email{wwzhu@tsinghua.edu.cn}
\affiliation{%
  \institution{Dept. of Computer Science and Technology, Tsinghua University}
}

\renewcommand{\shortauthors}{Chen Tang, et al.} 

\begin{abstract} 
Mixed-precision quantization (MPQ) suffers from the time-consuming process of searching the optimal bit-width allocation (\emph{i.e.,} the policy) for each layer, especially when using large-scale datasets such as ISLVRC-2012. This limits the practicality of MPQ in real-world deployment scenarios. To address this issue, this paper proposes a novel method for efficiently searching for effective MPQ policies using a small proxy dataset instead of the large-scale dataset used for training the model. Deviating from the established norm of employing a consistent dataset for both model training and MPQ policy search stages, our approach, therefore, yields a substantial enhancement in the efficiency of MPQ exploration. Nonetheless, using discrepant datasets poses challenges in searching for a transferable MPQ policy. Driven by the observation that quantization noise of sub-optimal policy exerts a detrimental influence on the discriminability of feature representations---manifesting as diminished class margins and ambiguous decision boundaries---our method aims to identify policies that uphold the discriminative nature of feature representations, \emph{i.e.,} intra-class compactness and inter-class separation. This general and dataset-independent property makes us search for the MPQ policy over a rather small-scale proxy dataset and then the policy can be directly used to quantize the model trained on a large-scale dataset. Our method offers several advantages, including high proxy data utilization, no excessive hyper-parameter tuning, and high searching efficiency. We search high-quality MPQ policies with the proxy dataset that has only 4\% of the data scale compared to the large-scale target dataset, achieving the same accuracy as searching directly on the latter, improving MPQ searching efficiency by up to 300$\times$. 
\end{abstract} 



\maketitle

\def\thefootnote{$\dagger$}\footnotetext{Corresponding authors}

\section{Introduction}
With the success of deep learning, deep neural networks (DNNs) have been adopted for many artificial intelligence tasks such as image classification \cite{he2016deep, howard2017mobilenets, sandler2018mobilenetv2}, object detection \cite{redmon2016you,ren2015faster}, and meanwhile, become the indispensable part of modern multimedia applications \cite{shi2019watch}. However, the large computational resource requirements of DNNs remain one of the most giant stumbling blocks for deploying deep learning models. There are several compression techniques to reduce the redundancy in a deep model, such as pruning \cite{liu2018rethinking}, knowledge distillation \cite{hinton2015distilling} and quantization \cite{choi2018pact,zhou2016dorefa,tang2022arbitrary, liu2023single}. Quantization is a promising technique to reduce both the storage and computational resources overhead remarkably, by leveraging the fact that the inference precision is not strictly as high as training time. Therefore, quantization enables large models to run directly on the edge and mobile devices without redesigning a new model architecture, which empowers edge intelligence significantly.

Quantization can be divided into two categories: fixed-precision quantization and mixed-precision quantization (MPQ). 
Fixed-precision quantization \cite{zhou2016dorefa,esser2020learned,liu2022nonuniform}, where an identical bit-width is designated for all layers in a deep model. 
While such a paradigm is proven to make the quantized model achieve sufficiently good performance at high bit-width (\emph{e.g.,} $\geq$ 8 bits), a uniform bit-width is challenging for quantization in an ultra-low bit-width (\emph{i.e.,} $\leq$ 4 bits) scenario. 
For example, BRQ \cite{han2021improving} reports that there is more than 20\% top-1 accuracy degradation in a 2 bit quantization for the MobilNetV2 model as compared to its full-precision (FP) counterpart. 

Mixed-precision quantization (MPQ) \cite{wang2019haq,guo2020single,huang2022sdq,elthakeb2020releq,yu2020search} offers a flexible and efficient way to quantize deep models by allocating varying bit-widths to individual layers based on their diverse redundancy levels. 
Unlike fixed-precision quantization, MPQ assigns specific precisions to different layers, with higher redundancy layers receiving less bit-width than lower redundancy ones, thereby achieving an optimal accuracy-efficiency trade-off. 
The MPQ process typically involves two stages: 
firstly, a full-precision (FP) model $\mathcal{M}_{FP}$ is trained on a training dataset $\mathcal{D}_{train}$; 
subsequently, the FP model is served as a weight initialization to be quantized, while simultaneously searching for the optimal MPQ policy for determining the quantization precision to each layer, over a searching dataset $\mathcal{D}_{search}$. 
The searching process also performs quantization-aware training, therefore all search-based approaches \cite{wang2019haq,cai2020rethinking,huang2022sdq} use the same dataset during both two processes, namely, $\mathcal{D}_{train} = \mathcal{D}_{search}$. 
Although using consistent datasets surely bring an accurate policy for the model to quantize, two problematic issues arise: 
\textbf{(a)} when $\mathcal{D}_{train} = \mathcal{D}_{search}$ and $\mathcal{D}_{train}$ is large-scale, the combinatorial nature \cite{tang2022mixed,wang2019haq} of the MPQ problem poses severe difficulties in search efficiency (\emph{e.g.,} BP-NAS consumes 35.6 GPU-hours to search for the ResNet-50 \cite{wang2021generalizable}). 
\textbf{(b)} in some sensitive user-data application scenarios, the training dataset is inaccessible. 

Nevertheless, few research has explored to decouple the dataset used in model training and MPQ search stages. This is promising to improve the search efficiency since the searching process can be done on a small-scale proxy dataset, but inevitably encounters intractable challenges due to shifted data distributions and reduced data volume caused by the disparate datasets. 
Notably, when CIFAR-10 was used to search for MPQ policy for ResNet50 trained on ISLVRC-2012, EdMIPS \cite{cai2020rethinking} encountered a substantial loss of nearly 7\% in Top-1 accuracy \cite{wang2021generalizable}. 
Recently, GMPQ \cite{wang2021generalizable} indicates that, for an input image, preserving the attribution rank between the FP and quantized model can search a generalizable MPQ policy. 
They resort the feature visualization technique Grad-cam \cite{selvaraju2017grad} to maintain the consistency of image attribution rank between the quantized and FP model. 
GMPQ can be regarded as an instance-level regularization over the proxy dataset, by enforcing a consistent relationship between FP and quantized model of each input instance. 
However, it is noteworthy that GMPQ does not harness information beyond the instance-level, namely, at the class-level. Furthermore, GMPQ entails intricate hyper-parameter tuning to align the attribution rank, contributing to its complexity. 

In this paper, we search the effective transferable MPQ policy by exploiting the class-level information on the proxy datasets, considering the class-level information is more luxuriant than instance-level \cite{chen2021generalized}. Our idea is motivated by the observation that quantization poses side effects to the quantized model in the feature space compared to the FP model. Our finding has shed light on a common drawback of quantization: the quantization noise remarkably narrows the margin between classes and blurs the decision boundary (see Figure \ref{fig_feature_vis}). On the other hand, maximizing inter-class separation while enhancing the intra-class compactness is highly favorable for classification, as there is a consensus that a large classification margin enhances the generalizability from statistical machine learning (\emph{e.g.,} SVM) to recent deep learning research \cite{wan2018rethinking,ranasinghe2021orthogonal}. We hence look for the MPQ policy that can properly gather the features of the same classes and separate the features of different classes, making the features more robust to quantization noise. 

Experimental results validate that a large margin between classes of proxy data helps search for a transferable MPQ policy for quantizing the model trained on challenging large-scale datasets. 
Our approach achieves competitive performance when searching on very small proxy datasets versus directly on large-scale datasets, in which the size of the former is only 4\% of the latter. 
Consequently, we improve the MPQ policy search efficiency impressively. 
For ResNet18 and MobileNetv1, by using StanfordCars \cite{KrauseStarkDengFei-Fei_3DRR2013} as the proxy dataset, our method achieves 375$\times$ and 300$\times$ speedup compared to the state-of-the-art MPQ approach FracBits \cite{yang2021fracbits}, respectively. 
\section{Related Work}

\subsection{Fixed-Precision Quantization} 
Fixed-precision quantization assigns a uniform bit-width for all layers. 
In this paper, we only consider quantization-aware training, as it can achieve higher compression ratio than post-training quantization \cite{nagel2020up,hubara2021accurate} and zero-shot quantization \cite{li2023hard,xu2023generative,yvinec2023powerquant}. 

Dorefa \cite{zhou2016dorefa} and PACT \cite{choi2018pact} uses a low-precision representation for weights and activations during forward propagation, and utilizes the Straight-Through Estimation (STE) \cite{bengio2013estimating} to estimate the gradient of piece-wise quantization function for backward propagation. 
To relieve the bias gradient of the STE, DSQ \cite{gong2019differentiable} employs tangent functions to approximate the non-differentiable quantization function. 
LSQ \cite{esser2020learned} introduces the learnable step-size scale factors to scale the tensor-wise weight and activation distributions. 
BSQ \cite{han2021improving} further applies a bin regularization to ensure the weights fall in the center of quantization bins. 
All these works focus on training a well-performing quantized network, but suffer from severe performance degradation when the bit-width is decreased significantly. 

\subsection{Mixed-Precision Quantization} 
The fundamental of Mixed-precision quantization (MPQ) is that the different layers in a model have different redundancy, in which the high redundancy layers can be allocated small bit-width to ensure low complexity without a severe performance drop. 
However, the bit-width choice is discrete, and the combination of bit-width and layer (\emph{i.e.,} the policy) grows exponentially. 
Therefore, the main challenge is how to determine the optimal bit-width for each layer. 

Obviously, brute-force is rather ineffective for the purpose of searching, as an $L$ layers model with $n$ bit-widths for activations and weights has $n^{2L}$ possible policies \cite{wang2019haq}. 
To solve this, several studies make efforts to apply the intelligent algorithms to search the optimal MPQ policy. 
HAQ \cite{wang2019haq} and ReleQ \cite{elthakeb2020releq} use reinforcement learning (RL) to train a bit-width allocator. 
SPOS \cite{guo2020single}, EdMIPS \cite{cai2020rethinking} and BP-NAS \cite{yu2020search} adopt neural architecture search (NAS) methods to learn the bit-width. 
In particular, GMPQ \cite{wang2021generalizable} develops an instance-level regularization to make searching MPQ policy on a small dataset possible. However, GMPQ suffers from a fussy hyper-parameters tuning, including the approximated attribution rank level, number of interested pixels, etc.

Unlike learning the optimal MPQ policy, HAWQ \cite{dong2019hawq,dong2020hawq} and MPQCO \cite{chen2021towards} use the Hessian information as the quantization sensitivity metrics to assist bit-width assignment. 
LIMPQ \cite{tang2022mixed} proposes to learn the layer-wise importance during a once quantization-aware training process. 
In contrast to these methods that aim to define some metrics to estimate the quantization sensitivity of layers, we propose to directly learn the effective bit-width configurations on a small proxy dataset. 
\begin{figure*}[htbp]
\includegraphics[width=\linewidth,scale=0.8]{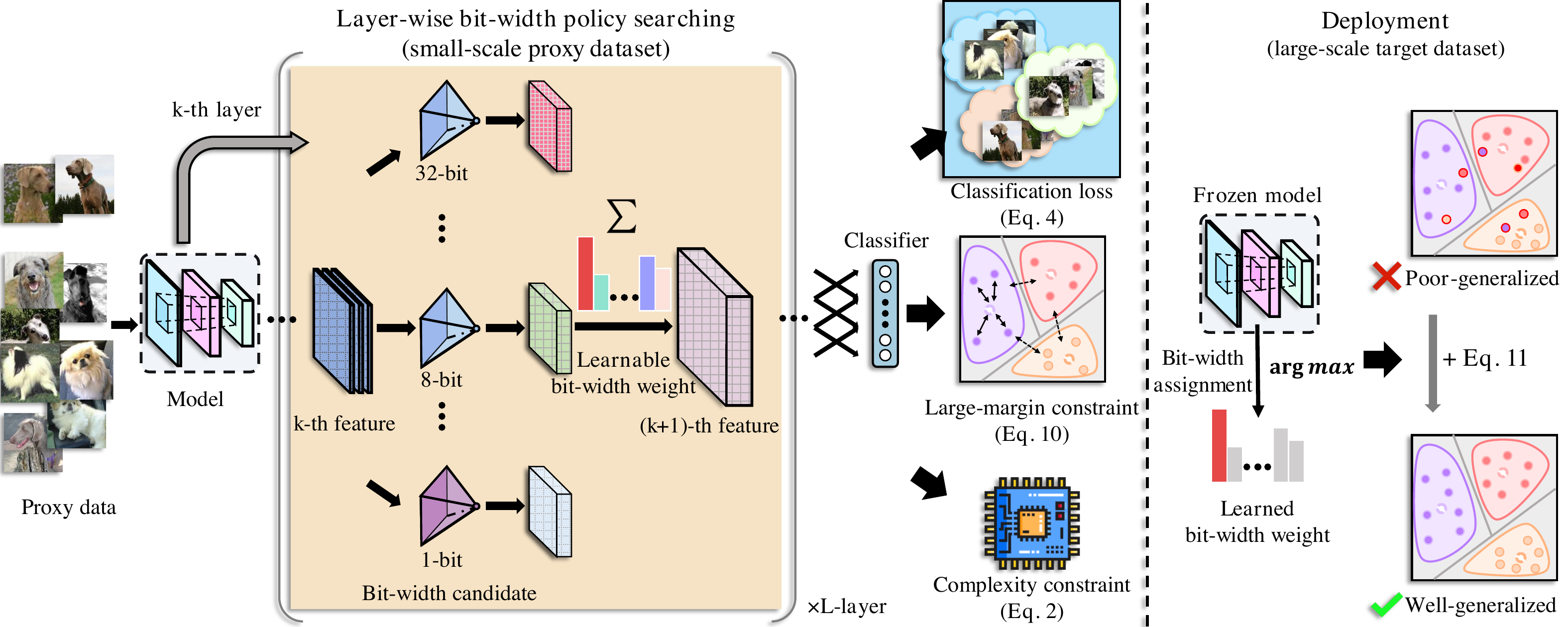}
\caption{
The illustration of our approach.
During the MPQ policy search process on the small-scale proxy dataset, we not only use the conventional classification loss and complexity loss as the optimization objective, but also introduce a large-margin constraint to search the policy can ensure the discriminative property in the feature space. 
In short, we hope the searched MPQ policy with a general and favorable attribute--gathering the features of the same classes and separating the features of different classes--to be applied to the target large-scale dataset (\emph{e.g.,} ISLVRC-2012) for model deployment efficaciously. 
} 
\label{_fig_illustration_framework}
\end{figure*}
\subsection{Discriminative Feature Learning}
Learning discriminative feature is highly favorable since it greatly facilitates the generalization of deep models, its core is to clarify the decision boundaries between classes. 
For nearly two decades, there are several studies to make efforts to achieve this. 

DrLIM \cite{hadsell2006dimensionality} proposes to use the contrastive loss to identify the classes. 
L-Softmax \cite{liu2016large} introduces a multiplicative hyper-parameter for the softmax function to produce a rigorous decision margin. 
L-GM \cite{wan2018rethinking} assumes the output of the penultimate layer (\emph{i.e.,} the deep features) follows the Gaussian Mixture (GM) distribution, and leverages the non-negative squared Mahalanobis distance to construct a GM loss.
OPL \cite{ranasinghe2021orthogonal} observes a potential orthogonality for features in the cross-entropy loss, and leverages this observation to explicitly enforce orthogonality of features. 
These works successfully demonstrate the significance of producing clear decision boundaries in the feature space, as the learned features become more robust and even increase the separation of features for the novel classes in a few-shot learning setting \cite{ranasinghe2021orthogonal}. 
\section{Method}
In this section, we first review the mixed-precision quantization (MPQ) problem in a differentiable way and discuss why it cannot be adopted on inconsistent datasets directly.
Next, we consider the MPQ policy searching from the feature perspective. 
Namely, what good MPQ policy can ensure the quantized model has a generalization deep feature as its full-precision counterpart? 
Motivated by the observation, we introduce the separation regularization to search the policy that guarantees the \emph{discriminative property of deep features}. The illustration of our approach is shown in Figure \ref{_fig_illustration_framework}. 

\subsection{Problem Formulation}
We consider a differentiable MPQ policy searching process \cite{cai2020rethinking,yu2020search,wang2021generalizable}.
Typically, the whole searching pipeline is organized as a Directed Acyclic Graph (DAG), where the nodes represent a specific quantization precision (\emph{e.g.,} 3bit), and the edges represent the learnable weight for its corresponding quantization precision. 
Therefore, a differentiable searching graph is built to determine the optimal quantization bit-width through the learnable weight, by adding a complexity constraint (\emph{e.g.,} BitOPs, model size) to the loss function. 

Accordingly, the loss function is defined as
\begin{equation} 
\mathcal{L}=\mathcal{L}_{task}+ \gamma \mathcal{L}_{comp},
\label{eq_original_searching_loss}
\end{equation}
where the $\mathcal{L}_{task}$ represents the task loss, \emph{i.e.,} the cross-entropy loss, that guarantees the classification accuracy, $\mathcal{L}_{comp}$ denotes the complexity loss that guarantees the target computational budget (\emph{i.e.,} BitOPs), and $\gamma$ is the hyper-parameters to control the accuracy-complexity trade-off. $\mathcal{L}_{comp}$ is defined as 
\begin{align}
&\mathcal{L}_{comp}=\sum_{l=0}^{L}\left(\sum_{j=0}^{||\bm{B^w}||}(p_j^{l, w} b_j^w) \sum_{k=0}^{||\bm{B^a}||}(p_k^{l, a} b_k^a) \right) comp^l, \\
&
\mbox{where} \quad p_j^{l, w} = \frac{\text{exp}({\alpha_j^l})}{\sum_{k=0}^{ ||\bm{B^w}||} \text{exp}({\alpha_k^l})} \quad p_k^{l, a} = \frac{\text{exp}({\beta_k^l})}{\sum_{k=0}^{||\bm{B^a}||} \text{exp}({\beta_k^l})}, \nonumber
\end{align}
$\bm{B^w}$ and $\bm{B^a}$ are the pre-defined bit-width candidate set for weights and activations, $\bm{\alpha^l}$ and $\bm{\beta^l}$ are the learnable weights vector for their corresponding bit-width candidate of layer $l$, \emph{e.g,} $\alpha^l_j \in \bm{\alpha^l}$ represents the learned weight for bit-width candidate $b^w_j \in \bm{B^w}$. 
$comp^l$ is the BitOPs constraint of layer $l$,
\begin{equation} 
comp^l= c_{in}^l \times c_{out}^l \times k_{a}^l \times k_{b}^l \times h_{out}^l \times w_{out}^l, 
\end{equation}
where $c_{in}$ and $c_{out}$ is the number of input and output channels, respectively. $k_a$ and $k_b$ are the kernel size, $w_{out}$ and $h_{out}$ are the width and height of the output feature map.
After searching, the bit-width for weights and activations of layer $l$ is determined by an $argmax$ function acts on its learnable weights vector $\bm{\alpha^l}$ and $\bm{\beta^l}$.

This paradigm and its variants \cite{cai2020rethinking,huang2022sdq} require the searching dataset to be consistent with the full-precision model training one, otherwise resulting in a serious accuracy degradation \cite{wang2021generalizable}. 
Inevitably, using a consistent dataset leads to inefficiencies, especially on large-scale datasets like ISLVRC2012 \cite{deng2009imagenet} with over 1 million samples to search for. 

However, when searching an MPQ policy on a proxy dataset (\emph{e.g.,} a small-scale dataset CIFAR-10 with only 50000 training samples) through Equation \ref{eq_original_searching_loss} and then directly applying it to the model trained on a large-scale dataset (\emph{e.g.,} ISLVRC2012), while the accuracy and complexity are both met, the accuracy on the proxy dataset is not of direct interest to us, because high accuracy on proxy dataset does not imply equivalent high accuracy on challenging large-scale datasets. 
One may argue that we can abridge the size of the target dataset to improve the efficiency, such as using a subset of target datasets to conduct MPQ search, but this would also result in serious performance degradation, as shown in Sec. \ref{_exp_ablation_tudy}.

Accordingly, instead of optimizing the above improper objective on the proxy dataset, we aim to search an MPQ policy that guarantees a large-margin on the proxy dataset to handle the incoming classes of the large-scale dataset. 
\subsection{Exploiting the Class-level Information}
\begin{figure}[t]
\hfill
\subfigure[\small  Full-precision]{\includegraphics[width=2.6cm]{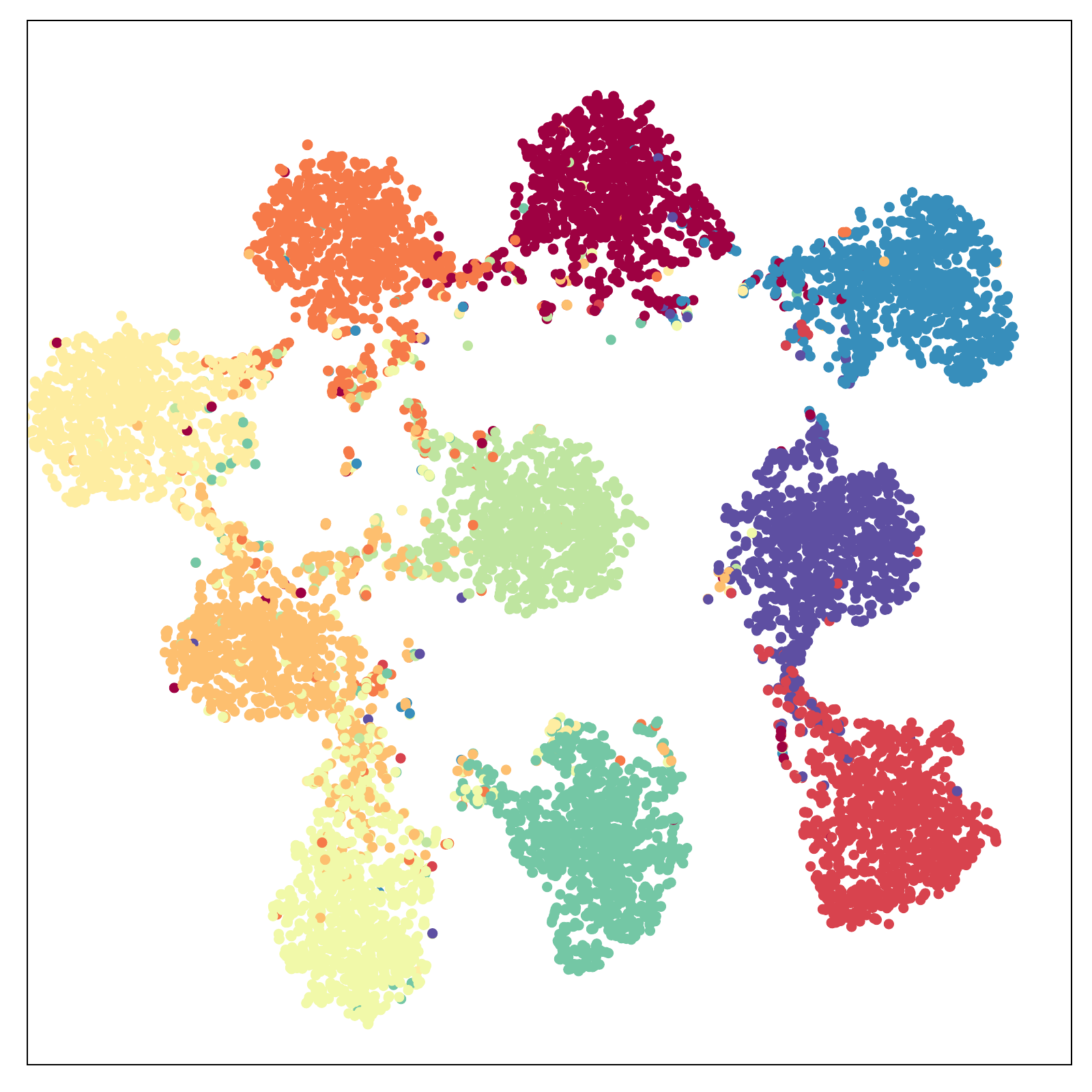}}
\hfill
\subfigure[\small Quantized]
{\includegraphics[width=2.6cm]{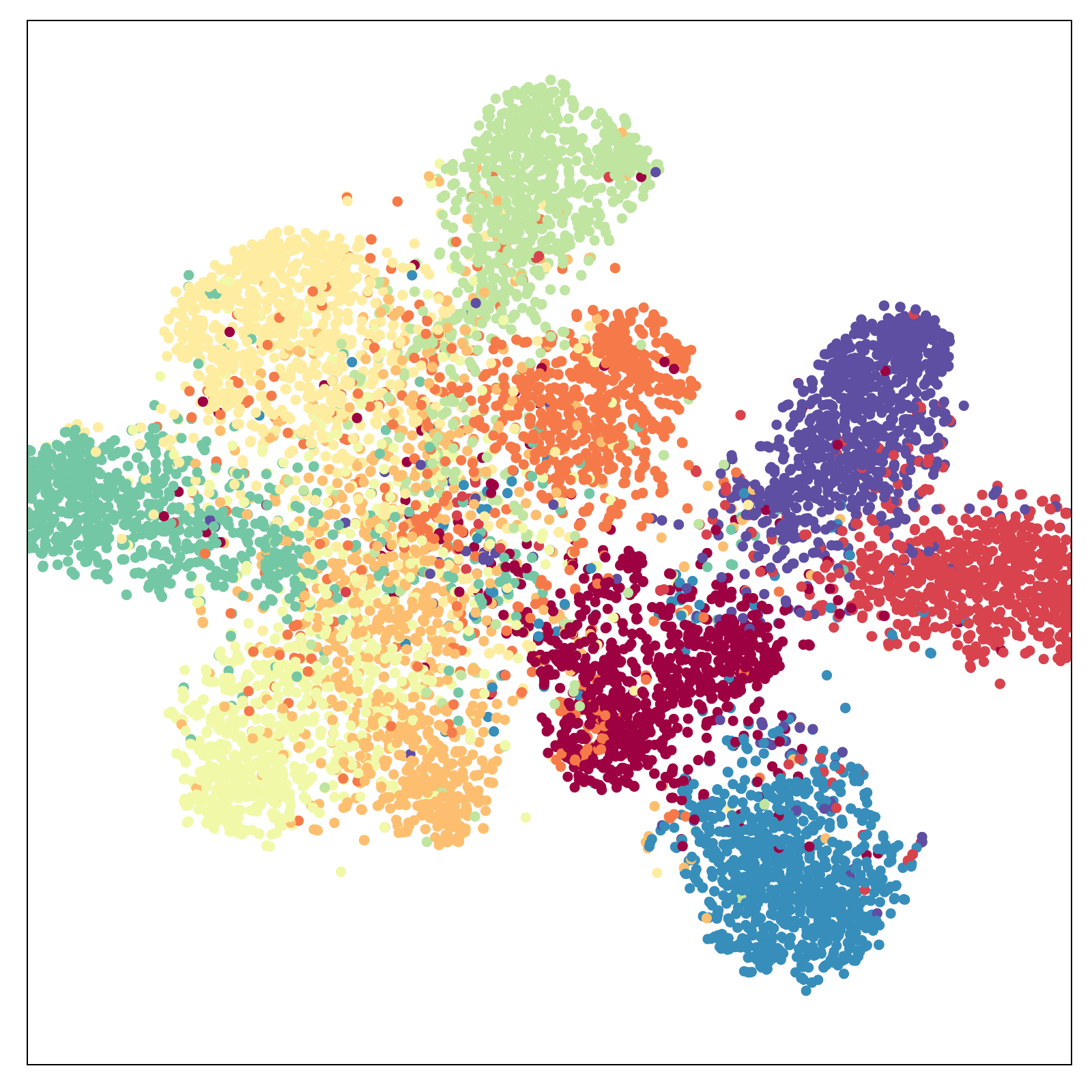}}
\hfill
\subfigure[Quantized (ours)]{\includegraphics[width=2.6cm]
{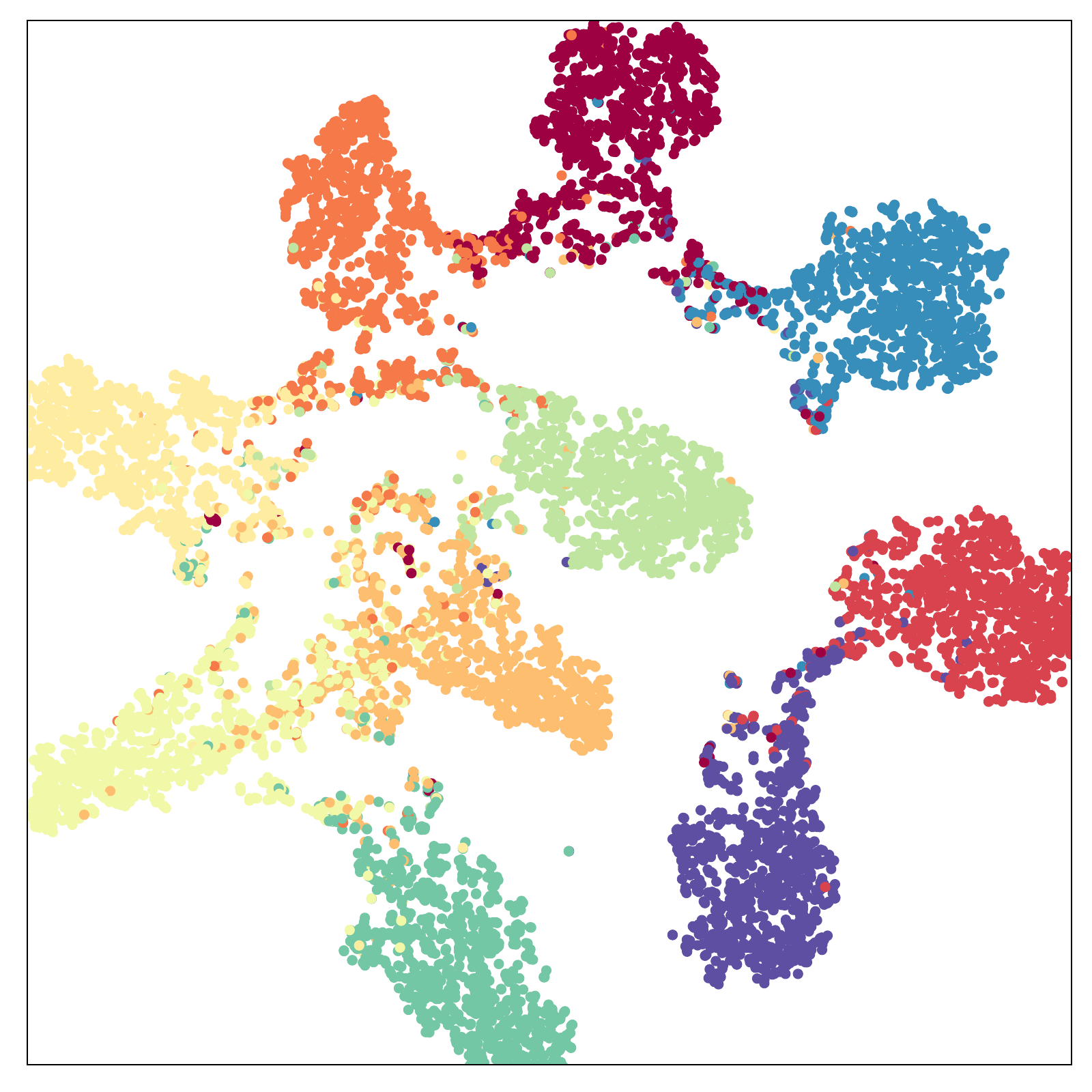}}
\hfill
\caption{The deep feature 2D visualization (t-SNE \cite{van2008visualizing}) on a proxy dataset CIFAR-10 over 
\textbf{(a)} the full-precision ResNet18,  
\textbf{(b)} direct searched MPQ policy through EdMIPS \cite{cai2020rethinking} and
\textbf{(c)} searched MPQ policy through proposed method. 
Colors represent different classes. 
}
\vspace{-0.5cm}
\label{fig_feature_vis}
\end{figure}

From the perspective of class-level features in a well-preforming MPQ policy, they should be well separated if not in the same class, and tightly gathered if in the same class. 
This has the following benefits: 
\textbf{a)} 
It alleviates the side effect of quantization on classification boundary.
As shown in Figure \ref{fig_feature_vis}(a) and Figure \ref{fig_feature_vis}(b), we observe quantization sharply narrows the class boundaries in the feature space compared to the full-precision model. 
Therefore, an MPQ policy with an explicit feature separation guarantee can effectivity alleviate the side effect of quantization. 
\textbf{b)} 
This is a widely pursued and \emph{dataset-independent} attribute, as from classical statistical machine learning to recent deep learning research \cite{wan2018rethinking,ranasinghe2021orthogonal,liu2016large} both recognize a large classification margin in feature space can help generalization. 

Motivated by this, we aim to search the MPQ policy that guarantees the large class margin on the proxy data distribution as much as possible. 
As we discussed above, such a general property in searched MPQ policies can ensure usability across the data distributions. 
However, the cross-entropy cannot provide this property, as the class margin is not explicitly formulated.
Therefore, the objective is not only to optimize accuracy and complexity, but also to find an MPQ policy that maximizes the class margin. 

We regard our approach as a class-level proxy data utilization, as it discovers the effective MPQ policy by leveraging the inter-class and intra-class information on the proxy dataset. 
The 2D visualization of our approach is shown in Figure \ref{fig_feature_vis}(c), we observe that the t-SNE pattern is quite similar to the full-precision model, indicating an MPQ policy that is able to separate the features is searched for the quantized model. 

\subsubsection{Separation Regularization} 
The first term in Equation \ref{eq_original_searching_loss} is the soft-max cross-entropy loss \cite{cai2020rethinking,wang2021generalizable}. 
For simplicity, we revisit it here by considering a binary classification problem, which can be trivial generalized to multi-class classification, 
\begin{align}
\mathcal{L}_{task} &= -\log \, \frac{\mbox{exp($\bm{w_1^{\intercal}} \bm{g}$)}}{\mbox{exp($\bm{w_1^{\intercal}} \bm{g}$)} + \mbox{exp($\bm{w_2^{\intercal}} \bm{g}$)}} \nonumber \\
&= -\log \, \frac{1}{\mbox{1} + \underbrace{ \mbox{exp($\bm{w_2^{\intercal}} \bm{g} - \bm{w_1^{\intercal}} \bm{g}$)}}_{\mbox{equivalent optimized term}}}, 
\label{_eq_ce_revisiting}
\end{align} 
where $\bm{w_1^{\intercal}}$ and $\bm{w_2^{\intercal}}$ are the weights for class 1 and class 2, respectively. $\bm{g}$ is the deep feature of the model produced by several convolution layers (\emph{i.e.,} layers that need to be quantized to mixed-precision).

Since the equivalent optimized term is not carried the margin objective during optimization, Equation \ref{_eq_ce_revisiting} cannot explicitly guarantee any margin between classes. 
Some previous works even observe that the learned feature regions for some classes tend to be bigger than others. 
If this combines with the side effect of quantization on decision boundaries, it inevitably leads to the search for sub-optimal MPQ policies.
In other words, the performance objective in Equation \ref{eq_original_searching_loss}, the cross-entropy, is improper when the MPQ searching and full-precision model training datasets are inconsistent. 

To this end, we introduce separation regularization to enforce a large margin guarantee in the searched policy.
Firstly, a small intra-class variance should be achieved to compact the features,
\begin{equation} 
\min\limits_{q} \sum_{i=1}^N q_i, \quad \mbox{where} \quad q_i=d(\bm{g_i}, \bm{\mu_{y_i}}),
\label{_eq_intra_class}
\end{equation}
where $N$ is the number of samples, $\bm{g_i}$, ${y_i}$ and $\bm{\mu_{y_i}}$ are the feature and label (ground truth) of sample $i$ and the feature mean of class $y_i$, respectively. $d(\cdot,\cdot)$ is the metric for calculating the distance between the feature and its mean (\emph{e.g.,} L2 distance). 

Secondly, we consider the inter-class margin by minimizing a classification loss as 
\begin{gather*} 
\min\limits \mathcal{L}_{cls} = \min\limits_{o} \sum_{i=1}^N \sum_{j=1}^K
o_{i,j}, \quad
\end{gather*}
\begin{gather}
o_{i,j}= 
\begin{cases}
-\log \frac{\exp (h_j(\bm{g_i}; m))}{{\sum\limits_{k=1}^K \mathbbm{1}(k \neq y_i) \exp \mbox{($h_k(\bm{g_i}; 0)$)} + \exp \mbox{($h_j(\bm{g_i}; m)$)}}, }, & \text{ if $ j = y_i $ } \\
0, & \text{ otherwise, }
\end{cases}
\label{_eq_margin_loss}
\end{gather}
$h(\cdot; \cdot)$ is a map from feature space $\mathbbm{R}^D$ (\emph{i.e.,} $\bm{g}$) to class-wise prediction score and will be introduced in Equation \ref{_eq:class_magin_GDM}. $\mathbbm{1}(\cdot)$ is the indicator function and $K$ represents the number of classes, respectively. 
$m$ is a non-negative scalar that represents the margin of different classes to form an explicit classification margin between the label class of sample $i$ and other classes in feature space, \emph{i.e.,} $h_j(\bm{g_{i}}; m)>h_k(\bm{g_{i}}; 0)$ $(k \neq j \mbox{, and } j = {y_i})$. 
One can see Equation \ref{_eq_margin_loss} becomes the classic log-softmax cross-entropy loss when $h(;m)$ is a linear transformation and $m \equiv 0$, 
\emph{e.g.,} in classic softmax cross-entropy, a linear layer with weight $\textbf{W} \in \mathbbm{R}^{D \times K}$ and no biases is used to project the deep feature $\bm{g_i}$ to $\mathbbm{R}^K$--let us denote $\bm{w_j}$ is the $j$-th column vector of $\textbf{W}$, thus $h_j(\bm{g_i}; 0) = \bm{w_j^{\intercal}} \bm{g_i}$. 
Please note when $m \neq 0$, the classification margin requires the output sign of $h$ should be always either positive or negative, which is not always satisfied in a classic softmax cross-entropy loss as the sign of linear projection is not certain. 

We hence follow the previous work L-GM \cite{wan2018rethinking} that assumes the feature $\bm{g_i}$ follows a Gaussian Mixture Distribution (GMD). 
Namely,
\begin{equation} 
p(\bm{g_i}) = \sum_{k=1}^K p(k) \mathcal{N}(\bm{g_i}; \bm{\mu_k},\, \bm{\Sigma_k}), 
\label{_eq_GMD}
\end{equation}
where $p(k)$ is the prior probability of class $k$, and $\bm{\mu_k}$ and $\bm{\Sigma_k}$ are the mean and covariance of class $k$.
The posterior probability of feature $\bm{g_i}$ is derived through the Bayes' rule,
\begin{align}
p(y_i|\bm{g_i}) &= \frac{p(y_i) \mathcal{N}(\bm{g_i}; \bm{\mu_{y_i}},\,\bm{\Sigma_{y_i}})} {p(\bm{g_i})} \nonumber \\ 
&= \frac{p(y_i)\mathcal{N}(\bm{g_i}; \bm{\mu_{y_i}},\,\bm{\Sigma_{y_i}})}{\sum_{k=1}^K p(k) \mathcal{N}(\bm{g_i}; \bm{\mu_k},\,\bm{\Sigma_k})}.
\end{align} 
Under the GMD assumption, we can easily derive the additive inter-class margin according to
\begin{align}
& h_{y_i}(\bm{g_i}; m) = p(y_i)\mathcal{N}(\bm{g_i}; \bm{\mu_{y_i}},\bm{\Sigma_{y_i}}, m) =  \notag \\
& p(y_i) | \bm{\Sigma_{y_i}}|^{-\frac{1}{2}} \exp \{-\left(\underbrace{\frac{1}{2}(\bm{g_i}-\bm{\mu_{y_i}})^{\intercal} \bm{\Sigma_{y_i}}^{-1} (\bm{g_i}- \bm{\mu_{y_i}})}_{\mbox{non-negative}}+m\right)\},
\label{_eq:class_magin_GDM} 
\end{align} 
where $h(\cdot)$ is formulated from a probability perspective, thus it is guaranteed to be non-negative. 
By replacing the subscript $y_i$ of Equation \ref{_eq:class_magin_GDM} with $k$ and setting $m=0$, we can derive
$h_{k}(\bm{g_i}; 0) = p(k)\mathcal{N}(\bm{g_i}; \bm{\mu_{k}},\bm{\Sigma_{k}}, 0)$. 
Substitute it and Equation \ref{_eq:class_magin_GDM} into Equation \ref{_eq_margin_loss}, we can obtain the $\mathcal{L}_{cls}$ accordingly. Finally, we apply a log-likelihood term \cite{wan2018rethinking} to restrict the feature $\bm{g_i}$ centralization near its mean $\bm{\mu_{y_i}}$ to achieve intra-class compactness according to Equation \ref{_eq_intra_class} and Equation \ref{_eq_GMD}, 
\begin{align} 
\mathcal{L}_{inc} &= \sum_{i=1}^N q_i = \sum_{i=1}^N d(\bm{g_i}, \bm{\mu_{y_i}}) \nonumber \\
&= \sum_{i=1}^N -\log \mbox{ $p(y_i) \mathcal{N}(\bm{g_i}; \bm{\mu_{y_i}},\,\bm{\Sigma_{y_i}})$}.
\end{align} 
We assume $p(y_i)=\frac{1}{K}$ and $\bm{\Sigma_{y_i}}$ is diagonal for both simplicity and considering its application in existing research  \cite{dudoit2002comparison,wan2022shaping}. 

Thus, the optimization objective during MPQ searching is
\begin{equation} 
\mathcal{L}=\mathcal{L}_{cls}+ \lambda \mathcal{L}_{inc} + \gamma \mathcal{L}_{comp}, 
\end{equation}
where $\mathcal{L}_{cls}$ is the classification loss, $\mathcal{L}_{inc}$ is the intra-class compactness loss and $\mathcal{L}_{comp}$ is the complexity loss. $\lambda$ and $\gamma$ are the hyper-parameters to weight the corresponding loss in the optimization process. 
\section{Experiment}
\subsection{Settings}
\subsubsection{Datasets}
The proxy (MPQ policy searching) datasets $\mathcal{D}_{search}$ are CIFAR-10 \cite{krizhevsky2009learning} and StanfordCars \cite{KrauseStarkDengFei-Fei_3DRR2013}.
CIFAR-10 has 10 categories, and each category has 5000 training samples and 1000 test samples. 
StanfordCars has 196 categories of cars; and the training set has 8144 training samples, and the test set has 8041 test samples. 
The target (model training) dataset $\mathcal{D}_{train}$ is ISLVRC-2012 \cite{deng2009imagenet} with 1000 categories, containing about 1.28M training samples and 50000 validation samples. 

We search the MPQ policy on the training set of proxy datasets. 
The training samples of proxy datasets are used to search MPQ policies.
After searching, we finetune (quantize) the model with the searched policies on the target dataset.  
We use the basic data augmentation methods during finetuning and evaluate the final performance on the ISLVRC-2012 validation set. 

\subsubsection{Models}
We conduct the experiments on three representative models including the ResNet-\{18, 50\} \cite{he2016deep} and the MobileNet \cite{howard2017mobilenets}.
Particularly, we use the standard architecture for ResNet. 

\subsubsection{Hyper-parameters} 
For ResNet and MobileNet, the bit-width candidates of weights and activations are $\bm{B^w} = \bm{B^a} = \{2,3,4,6\}$ and $\bm{B^w} = \bm{B^a} = \{2,3,4,5,6\}$, respectively. 
Following the previous arts \cite{wang2019haq,esser2020learned,tang2022mixed}, the first and last layers are fixed to 8 bits. 

For searching, we adopt the SGD optimizer, and the initial learning rate is set to $0.01$ for 15 epochs. 
Empirically, we find the intra-class compactness regularization is not sensitive to the hyper-parameter and set $\lambda=0.1$ for all proxy datasets, more details for $\lambda=0.1$ can be found in the ablation study. 
We set the class margin $m=0.3$ and $m=0.01$ for CIFAR-10 and StanfordCars respectively while multiplying by the non-negative term in Equation \ref{_eq:class_magin_GDM}. 
We fine-tune the hyperparameter $\gamma$ in line with prior works on differentiable MPQ \cite{cai2020rethinking,wang2021generalizable}. A higher $\gamma$ value corresponds to a less computation complexity policy to search for. 

For finetuning (quantizing), we follow the basic quantization-aware training settings in LSQ \cite{esser2020learned} and LIMPQ \cite{tang2022mixed}. 
Specifically, we use the full-precision model (trainined on $\mathcal{D}_{train}$) as the initialization and adopt the SDG optimizer with Nesterov momentum \cite{sutskever2013importance} and the initial learning rate and weight decay are set to $0.04$ and $2.5 \times 10^{-5}$, respectively. 
We use the cosine learning rate scheduler and finetune the model 90 epochs and the first 5 epochs are used as warm-up. 

\begin{table}[t!]
    \caption{Accuracy and efficiency results for ResNet. 
    ``Top-1 Q/FP'' represents the Top-1 accuracy of quantized model and full-precision model. 
    ``MP'' means mixed-precision quantization.
    ``Cost'' denotes the MPQ policy search time that is measured by GPU-hours. 
    ``*'': reproduces through the vanilla ResNet architecture \cite{he2016deep}. 
    ``\#'': the result of shortening the search epochs to half. 
    ``Ours-C'': denotes the MPQ policies search on CIFAR-10. 
    ``Ours-S'': denotes the MPQ policies search on StanfordCars. 
    The lowest accuracy degradation results are bolded in each metric. 
    }
    \setlength{\tabcolsep}{1.1mm}
    \centering
    \begin{tabular}{cccccc}
    \toprule
    Method     & W-bits    & A-bits     & Top-1 Q/FP (\%)   & BitOPs (G) & Cost (h) \\ \midrule
    \multicolumn{6}{c}{ResNet18}                                                    \\ \midrule
    PACT       & 3         & 3          & 68.1 / 70.4       & 23.09      & -        \\
    LSQ$^*$    & 3         & 3          & 69.4 / 70.5       & 23.09      & -        \\
    EdMIPS     & 3MP       & 3MP        & 68.2 / 69.6       & -          & 9.8      \\
    GMPQ$^*$   & 3MP       & 3MP        & 68.6 / 70.5       & 22.8       & 0.6      \\
    DNAS       & 3MP       & 3MP        & 68.7 / 71.0       & 25.38      & -        \\
    FracBits   & 3MP       & 3MP        & 69.4 / 70.2       & 22.93      & 150.1    \\
    LIMPQ      & 3MP       & 3MP        & 69.7 / 70.5       & 23.07      & 3.3       \\
    \rowcolor{gray!15}Ours-C     & 3MP       & 3MP        & \textbf{70.0} / \textbf{70.5} & 23.07 & 0.9\\
    \rowcolor{green!10}Ours-S     & 3MP       & 3MP        & 69.6 / 70.5 & 23.06 & 0.3\\
    \midrule
    PACT       & 4         & 4          & 69.2 /  70.4      & 35.04      & - \\
    LSQ$^*$    & 4         & 4          & 70.5 /  70.5      & 35.04      &  - \\
    DNAS       & 4MP       & 4MP        & 70.6 /  71.0      & -          & -  \\
    FracBits   & 4MP       & 4MP        & 70.6 /  70.2      & 34.7      & 151.3    \\
    FracBits\#  & 4MP       & 4MP        & 70.3 /  70.2      & 34.7      & 76.8    \\
    LIMPQ      & 4MP       & 4MP        & 70.8 /  70.5      & 35.04      & 3.3 \\
    \rowcolor{gray!15}Ours-C    & 4MP        & 4MP        &  \textbf{70.8} / \textbf{70.5} & 34.7 & 0.9 \\
    \rowcolor{green!10}Ours-S   & 4MP        & 4MP        &  70.5 / 70.5 & 34.7 & 0.4 \\\midrule
    \multicolumn{6}{c}{ResNet50}                                                    \\\midrule
    HAQ        & 4MP       & 8          & 76.1 /  76.2      & 136.5      & - \\
    BP-NAS     & 4MP       & 4MP        & 76.7 /  77.5      & 64.4       & 35.6  \\
    FracBits   & 4MP       & 4MP        & 76.5 /  77.5      & 71.17      & 630.6 \\
    \rowcolor{gray!15}Ours-C  & 4MP     & 4MP         & \textbf{76.8 / 77.5} & 70.43 & 1.73\\
    \rowcolor{green!10}Ours-S & 4MP     & 4MP         & 76.2 / 77.5   & 71.6 & 1.3 \\
    \bottomrule
    \end{tabular}
    \vspace{-0.5cm}
    \label{tab_resnet18_result}
\end{table}

\subsection{Comparisons with the State-of-the-Art}
\label{_exp_main_comparisons}
We compare our method with the SOTA quantization works on the classification task. 

For fixed-precision works, we compare our method with PACT \cite{choi2018pact}, PROFIT \cite{park2020profit} and LSQ \cite{esser2020learned}. 
For MPQ works, we compare our method with DNAS \cite{wu2018mixed}, 
HMQ \cite{habi2020hmq}, HAQ \cite{wang2019haq}, BP-NAS \cite{yu2020search}, FracBits \cite{yang2021fracbits}, GMPQ \cite{wang2021generalizable}, SDQ \cite{huang2022sdq} and LIMPQ \cite{tang2022mixed}. 

Specifically, since original LSQ and GMPQ use the Pre-Activation ResNet architecture, we re-implement them for fair comparisons through the vanilla ResNet \cite{he2016deep}.
\subsubsection{ResNet}
We show the mixed-3bits and mixed-4bits results of ResNet-\{18, 50\}, as listed in Table \ref{tab_resnet18_result}. 
We provide the full-precision accuracy to compare the \emph{absolute accuracy degradation} between the full-precision and quantized model. 

For ResNet18, under 3-bits level BitOPs constraints, ``Ours-C'' causes only $0.5\%$ Top-1 accuracy degradation compared to the full-precision model, which is the lowest one among recent works. 
Under 4bits level BitOPs constraints, ``Ours-C'' achieves the highest Top-1 accuracy. 
Meanwhile, it achieves about 160$\times$ policy search speedup compared with FracBits. 
Thanks to the small data amounts of StanfordCars,  ``Ours-S'' uses only 8041 training samples to search a very competitive MPQ policy. 

For ResNet50, we search 4bits level policies. One can see that our method achieves quite similar performance compared to gradient-based methods BP-NAS and FracBits while further reducing the search time significantly. 

Overall, our method not only achieves a comparable accuracy as searching directly on ISLVRC-2012, but also significantly improves the searching efficiency. 

\subsubsection{MobileNet}
Table \ref{tab_mobilenetv1_result} summarizes the results of mixed-3bits and mixed-4bits on MobileNetv1. 

\begin{table}[t] 
\setlength{\tabcolsep}{1.1mm}
\centering
\caption{Accuracy and efficiency results for MobileNetv1.
``Top-1/5'' represents Top-1 and top-5 accuracy respectively. 
}
\begin{tabular}{cccccc}
\toprule
Method   & W-bits    & A-bits     & Top-1/5 (\%)     & BitOPs (G) & Cost (h) \\ \midrule
PACT     & 4         & 4          & 62.4 / 82.2      & 9.68       & - \\
LSQ      & 3         & 3          & 68.3 / 88.1       & 5.8       & -\\
HMQ      & 3MP       & 4MP        & 69.3 /  -        & -          & -  \\
FracBits & 3MP       & 3MP        & 68.7 /  88.2     & 5.78       & 237.2 \\
LIMPQ    & 3MP       & 3MP        & 69.5 /  89.1     & 5.78       & 3.4 \\
\rowcolor{gray!15} Ours-C     & 3MP       & 3MP        &  \textbf{69.9}  / \textbf{89.3} & 6.28 & 1.0\\
\rowcolor{green!10}Ours-S     & 3MP       & 3MP        & 69.6 / 89.2 & 6.13 & 0.8\\
\midrule
PACT     & 6         & 4          & 67.5  /  87.8     & 14.13     & - \\
PROFIT   & 4         & 4          & 69.1  /  88.4     & 9.68      & - \\
LSQ      & 4         & 4          & 71.2  /  90.0     & 9.68      & - \\
HAQ      & 4MP       & 4MP        & 67.5  /  87.9     & -         & 35.6 \\
HAQ      & 6MP       & 4MP        & 70.4  /  89.7     & -         & - \\
FracBits & 4MP       & 4MP        & 71.4  /  90.0     & 9.63      & 250.2  \\
LIMPQ    & 4MP       & 4MP        & 71.8  /  90.4     & 9.68      & 3.6 \\
\rowcolor{gray!15} Ours-C  & 4MP        & 4MP        &  \textbf{71.8}  /  \textbf{90.5} & 9.30 & 1.1 \\ 
\rowcolor{green!10}Ours-S     & 4MP       & 4MP        & 71.7 / 90.3 & 9.86 & 0.8\\
\bottomrule
\end{tabular}
\vspace{-0.5cm}
\label{tab_mobilenetv1_result}
\end{table}

For mixed-3bits searched on CIFAR-10, we observe our method both outperforms the existing SOTA mixed-precision work LIMPQ and fixed-precision work LSQ.
In particular, our method arises a 1.8\% absolute gain on Top-1 accuracy compared to LSQ, and 1.2\% higher accuracy than FracBits. 
We further narrow the gap between the full-precision and quantized MobileNet. 
Please note that we are the first work to provide a 3-bits level MobileNet that almost achieves 70\% Top-1 accuracy. 
For mixed-4bits searched on CIFAR-10, our method has up to 237$\times$ searching efficiency improvement compared to FracBits and up to 0.4\% higher accuracy compared to the SOTA efficient MPQ approach LIMPQ. 

For mixed-3bits and mixed-4bits searched on StanfordCars, they show 0.3\% and 0.1\% absolute Top-1 accuracy degradation compared to the CIFAR-10 but further save about 20\% searching cost. 
This further proves that our method can still be very effective even if the proxy dataset (\emph{i.e.,} all cars) has much lower class-similarity to the target dataset. 

\subsubsection{Discussion for Proxy Datasets}
In this subsection, we observe that using CIFAR-10 as a proxy dataset can search for more well-performing MPQ policies better than StanfordCars. 
On the other hand, StanfordCars has higher search efficiency than CIFAR-10. 
We conjecture this is because the category of CIFAR-10 is more similar to the target dataset ISLVRC-2012, and the data amounts of CIFAR-10 are more than that of StanfordCars. 
Meanwhile, we find that the performance loss of policies searched on StanfordCars is slightly larger than CIFAR-10 when the complexity constraint becomes tighter, \emph{e.g.,} the mixed-3bits results for MobileNet. 

Therefore, while it is feasible to search a well-performing MPQ policy by using an arbitrary proxy dataset, if the model requires more aggressive quantization, a proxy dataset with more class-level similarity compared to the target dataset could be considered to further improve the performance. 

\subsection{Complexity-Accuracy Trade-off}
\begin{figure}[t]
    \includegraphics[width=\linewidth,scale=1.00]{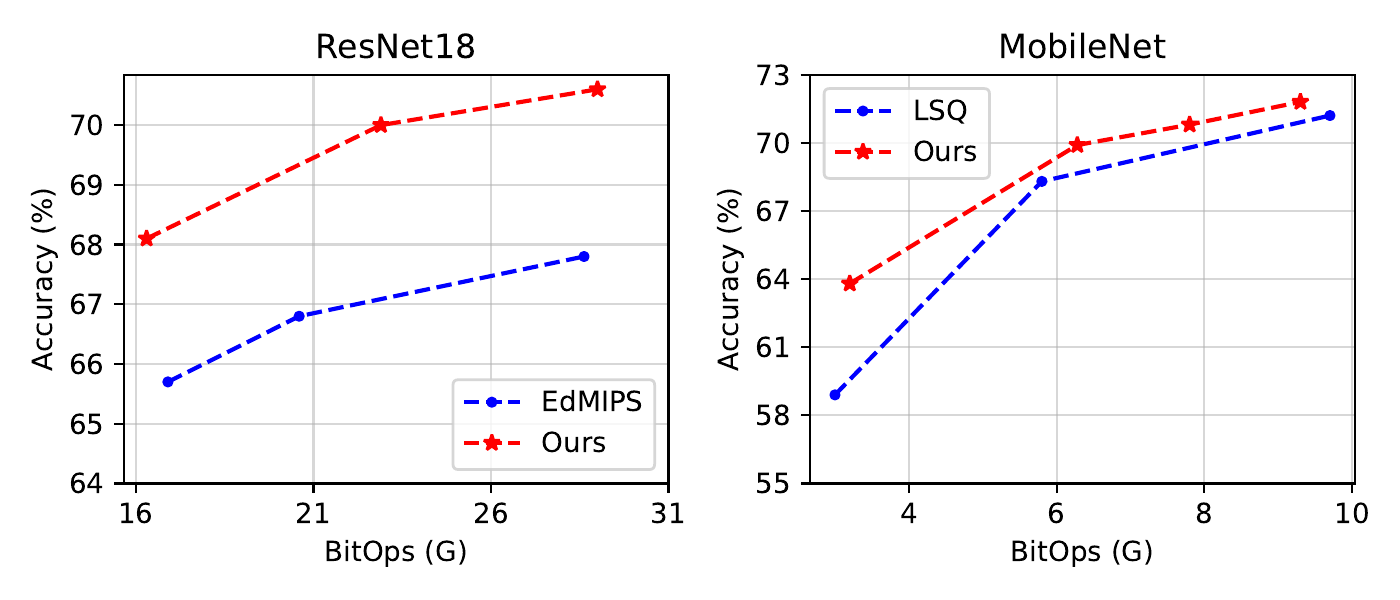}
    \caption{Complexity-accuracy trade-off for ResNet18 and MobileNet.}
    \label{_fig_resnet18_mbv1_complexity_accuracy_trade_off}
    \vspace{-0.3cm}
\end{figure}

In Figure \ref{_fig_resnet18_mbv1_complexity_accuracy_trade_off}, we show the complexity-accuracy trade-off of LSQ \cite{esser2020learned}, EdMIPS \cite{cai2020rethinking} and our method for ResNet18 and MobileNet. 
Unless otherwise specified, the proxy dataset used in our method is CIFAR-10. 

For ResNet18, our method achieves significant performance gains compared to the mixed-precision approach EdMIPS. 
We even consistently have an absolute advantage of over 2\% Top-1 accuracy. 

For MobileNet, our method provides a very high accuracy improvement within the constraints of approximate complexity. 
Especially, our method improves 4.9\% Top-1 accuracy compared to LSQ at 3G BitOPs constraint. 
Meanwhile, our method has a much fine-grained trade-off thanks to the mixed-precision quantization. 

\begin{figure*}[h]
\begin{center}
\subfigure[Bit-width assignment for weights.]{
\hspace{-0.2cm}
\includegraphics[height=2.8cm,width=3.7cm]{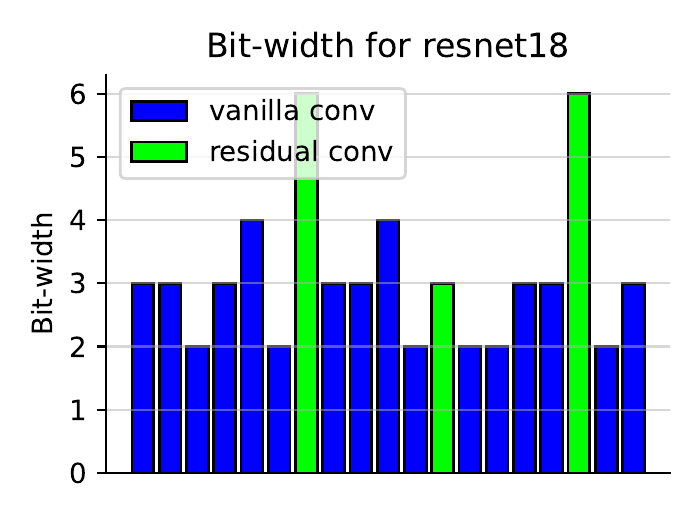} 
\hspace{-0.4cm}
\includegraphics[height=2.8cm,width=11.7cm]{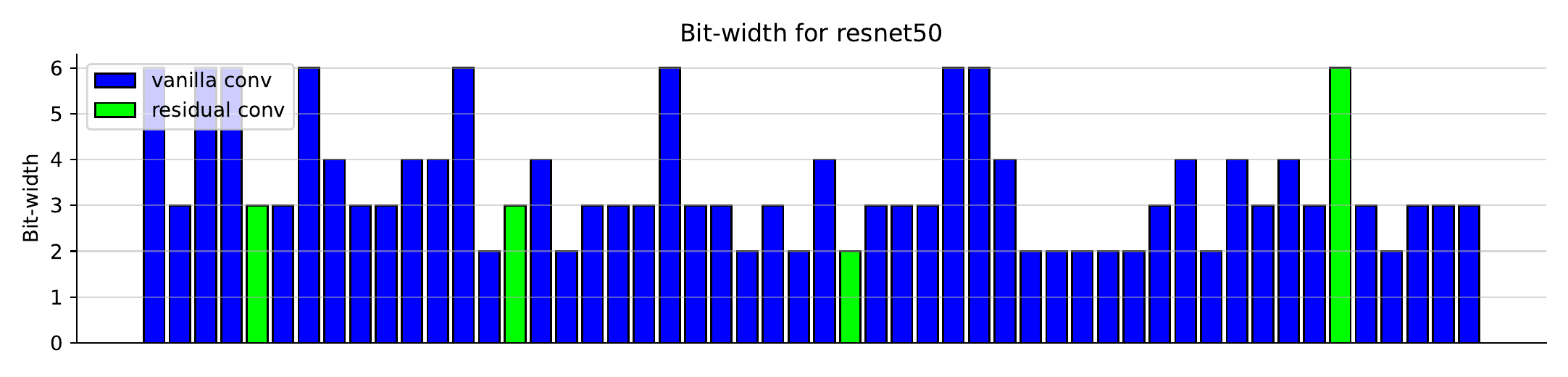} 
\hspace{-0.8cm}
\includegraphics[height=2.8cm,width=3.7cm]{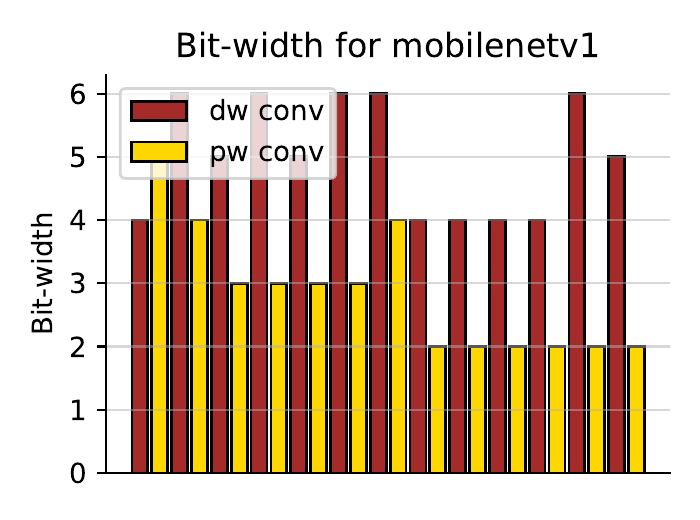}
}
\subfigure[Bit-width assignment for activations.]{
\hspace{-0.2cm}
\includegraphics[height=2.8cm,width=3.7cm]{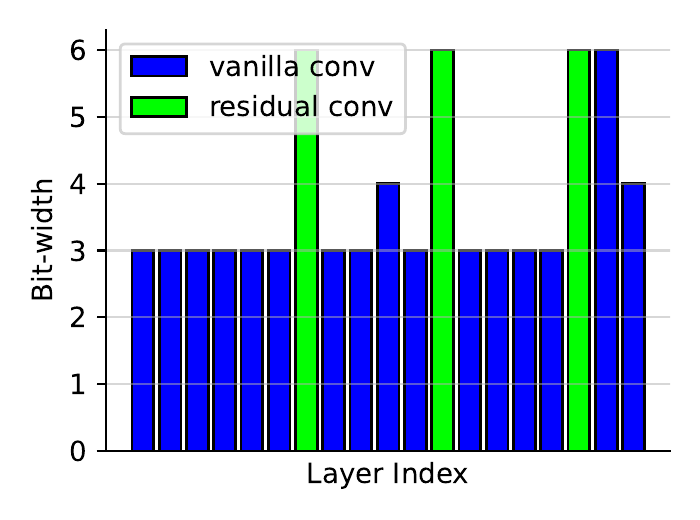
}
\hspace{-0.4cm}
\includegraphics[height=2.8cm,width=11.7cm]{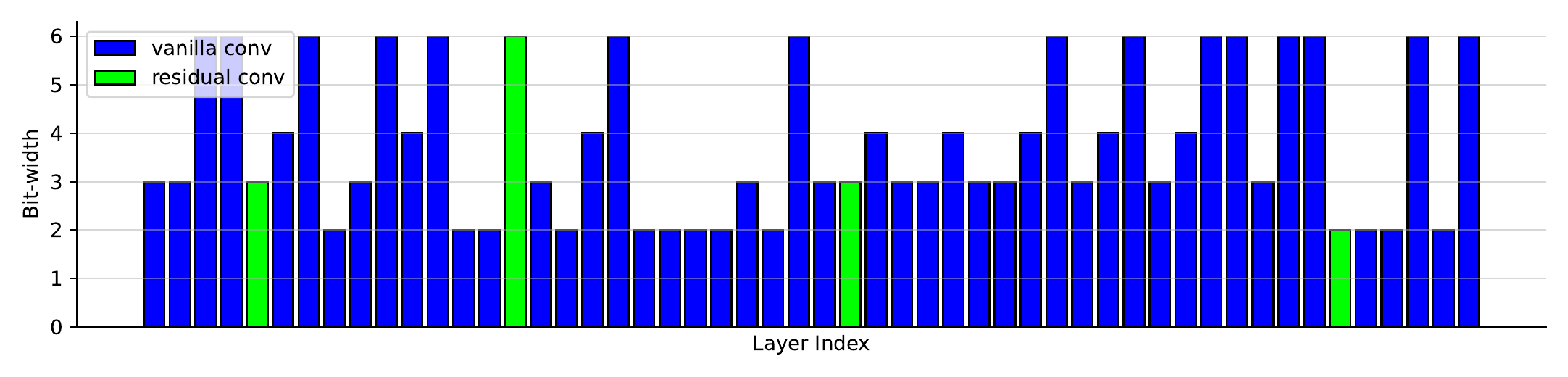}
\hspace{-0.8cm}
\includegraphics[height=2.8cm,width=3.7cm]{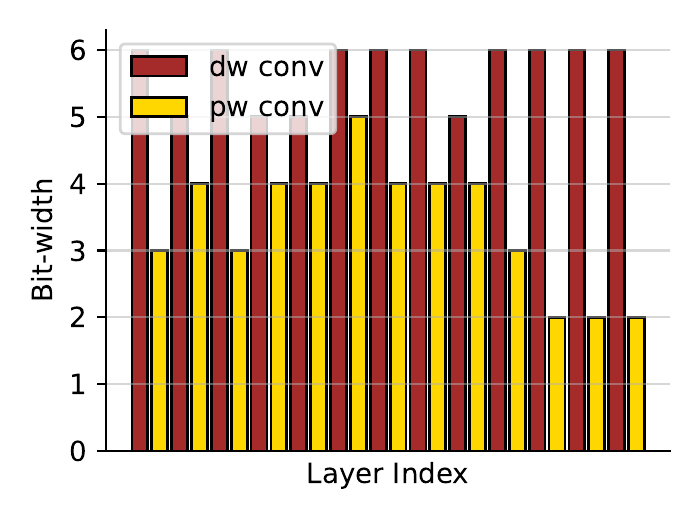}
}
\caption{Bit-width assignment for mixed-3bit ResNet18, ResNet50 and MobileNetv1.}
\vspace{-0.5cm}
\label{_fig_bitwidth_assignment_vis}
\end{center}
\end{figure*}

\subsection{Ablation Study} 
In this subsection, we investigate: 
(a) the effectiveness of using
a subset of $\mathcal{D}_{train}$ as the searching dataset; 
(b) what happens when one adds SEAM to the baseline; 
(c) performance difference under various hyper-parameters settings. 
\label{_exp_ablation_tudy} 
\subsubsection{Subset of ISLVRC-2012}
Although GMPQ has shown direct searching over the proxy dataset incurs severe performance degradation, there is no relevant literature to study the effect of using a subset of the target dataset (\emph{e.g.,} ISLVRC-2012) as the proxy dataset. 
To this end, we randomly sample 4\% (roughly the same sample size as CIFAR-10) training data from ISLVRC-2012 and use them to search a 3-bits level policy for ResNet18 without/with the proposed method. 

\begin{table}[t]
\caption{Results of ablation study. $\mathcal{D}_{search}$ denotes the dataset used for MPQ policy searching. ISLVRC-2012 (4\%) indicates a subset of ISLVRC-2012 with a sample of 4\% of the full training set. }
\begin{tabular}{ccc}
\toprule
Proxy ($\mathcal{D}_{search}$)      & SEAM            & Top-1 Quant (\%) \\ \midrule
ISLVRC-2012 (4\%)                   & \ding{56}       & 69.1      \\
ISLVRC-2012 (4\%)                   & \ding{52}       & 69.8      \\
CIFAR-10                            & \ding{56}       & 68.4      \\
CIFAR-10                            & \ding{52}       & 70.0      \\ \bottomrule
\end{tabular}
\label{tab_ablation_tudy}
\end{table}

As shown in Table \ref{tab_ablation_tudy}, the subset of ISLVRC-2012 without proposed method still has about 1\% performance degradation compared to CIFAR-10 with proposed method. 
This is because the data distribution in the subset is significantly different from the full set. 
When the proposed method is enabled, this subset yields superior performance than StanfordCars.
That further demonstrates the effectiveness of our method, and indicates that we can gain more performance by leveraging the class-similarity between proxy and target datasets. 

\subsubsection{Performance Improvement over Baseline}
\begin{table}[t]
    \centering
    \small
    \caption{Effectiveness of proposed method SEAM upon EdMIPS.}
    \begin{tabular}{cccccc}
    \toprule
    Method    & SEAM & $\mathcal{D}_{search}$ & Cost (h) & Top-1 (\%) & W\&A bits \\ \midrule
    EdMIPS    & \ding{56} & ISLVRC-2012  & \textbf{9.5}  & 65.9   & 2MP   \\ 
    EdMIPS    & \ding{52} & ISLVRC-2012  & 11.5   &  \textbf{66.4}  & 2MP  \\ 
    \bottomrule
    \end{tabular}
    \label{tab_ab_study_subs2}
\end{table}
To show that MPQ benefits from discriminability of feature representations, we further add proposed method on EdMIPS \cite{cai2020rethinking} -- a baseline MPQ approach. 
Specifically, EdMIPS is a conventional differentiable mixed-precision quantization approach, requiring consistency dataset of model training and policy searching. 
We directly apply proposed large-margin regularization term on it to search a MPQ policy. 
As shown in Figure \ref{tab_ab_study_subs2}, we observe that the proposed method can help the baseline to discover better MPQ policy.  

\subsubsection{Effectiveness of $\lambda$}
The setting of fixed $\lambda$ is inspired by several regularization-based quantization-aware training studies \cite{han2021improving,alizadeh2020gradient,chmiel2020robust}. 
Specifically, these studies enable their regularization term until after tens of training epochs. 
This delay is intended to \textbf{avoid optimization interference by the different loss terms}, ensuring that cross-entropy (CE) term dominates early training to optimize the parameters properly. 
Once the CE loss becomes small, the regularization term is added and plays a major role in optimization. 
In this paper, we direct use a small value for the regularization term $\lambda$ to simulate the above optimization idea. 
We ablate this hyper-parameter in Table \ref{tab_effectiveness_of_lambda}. 
\begin{table}[t]
\centering
\caption{Performance of different $\lambda$ values. }
\begin{tabular}{ccccc}
\toprule
Network & $\lambda$ & Top-1 (\%) & Cost (h) & W \&A bits \\ \midrule
ResNet18 &  0.02&  69.5  & 0.8 & 3MP  \\ 
ResNet18 & 0.1&  \textbf{70.0}  & 0.9 & 3MP  \\ 
ResNet18 & 0.5&  69.2  & 0.9 & 3MP  \\ 
ResNet18 &  2.5  & 69.0 & 0.9 & 3MP  \\ 
\bottomrule
\end{tabular}
\label{tab_effectiveness_of_lambda}
\vspace{-0.3cm}
\end{table}

One can see that $\lambda$ does need a relatively small value, which conforms with our optimization principle. 
\subsection{Bit-width Assignment Behavior} 
In Figure \ref{_fig_bitwidth_assignment_vis}, we visualize the searched MPQ policies for the mixed-3bit ResNet18, ResNet50 and MobileNet.
For ResNet, we clearly see that almost the highest bit-width is given for the residual convolution layers. 
That is because these layers are more important for bypassing signals from shallow to deep layers \cite{veit2016residual}, as well as having fewer parameters. 
For MobileNet, we find that higher bit-width is assigned to the Depthwise-Convolution (DW) layers than the Pointwise-Convolution (PW) layers, as the DW layer is typical less redundant \cite{tang2022mixed}. 

\subsection{Effectiveness of Knowledge Distillation} 
Follow SDQ \cite{huang2022sdq}, we use a ResNet101 as the FP distillation teacher during the fine-tuning time. The distillation temperature is set to 1. 
We compare our method with GMPQ and SDQ at the 3-bits levels (about 23G BitOPs) search policies. 

As shown in Tab \ref{tab_finetuning}, our approach achieves the highest performance when knowledge distillation is applied. 
In particular, compared to the state-of-the-art work SDQ under approximate complexity, our method attains an absolute accuracy improvement of 0.5\%, indicating our method can search the optimal MPQ policy properly on a small-scale proxy dataset for the purpose of knowledge distillation. 

\begin{table}[t]
\caption{Results of finetuning the ResNet18 with an external teacher model ResNet101 ($^{*}$: result from Table \ref{tab_resnet18_result}). 
}
\begin{tabular}{cccc}
\toprule
Method         & Teacher    & Top-1 (\%)          & BitOPs (G) \\ \midrule
Full-precision & -          & 70.5              & FP         \\
Ours - w/o KD$^{*}$  & -          & 70.0 (-0.5)     & 23.07      \\
GMPQ - KD   & ResNet101     & 69.5 (-1.0)     & 22.8       \\
SDQ - KD       & ResNet101  & 70.2 (-0.3)     & 23.5       \\ 
Ours - KD      & ResNet101  & 70.7 (\textbf{+0.2})     & 23.07      \\ 
\bottomrule
\end{tabular}
\label{tab_finetuning}
\vspace{-0.3cm}
\end{table}
\section{Conclusion}
In this work, we propose to search the MPQ policy on a small-scale proxy dataset for a model trained on a large-scale one. 
To bridge the inconsistent data distributions, we not only focus on optimizing the accuracy on the proxy dataset, but also enforce a large-margin of the searched MPQ policy should be met. 
We regard this as a \emph{class-level} data exploitation for the limited proxy data, which is more data efficient than the \emph{instance-level} data exploitation \cite{wang2021generalizable}. 
Our class-level data exploitation renders the search policies can compact the features in the same classes and separate the feature into different classes, which is a favorable and dataset-independent property. 
The experiments validate our idea, and we use only 4\% of data to search for the high quality MPQ policies, achieving the same accuracy as searching directly on the large-scale dataset, and speeding up the MPQ searching process by up to 300$\times$.

\section{Acknowledgment}
This work is supported in part by Shenzhen Science and Technology Program (Grant No. RCYX20200714114523079 and JCYJ2022081810-\\1014030). 
The authors would like to thank the anonymous reviewers for their valuable comments. 

\bibliographystyle{ACM-Reference-Format}
\balance
\bibliography{reference}


\begin{thebibliography}{53}


\ifx \showCODEN    \undefined \def \showCODEN     #1{\unskip}     \fi
\ifx \showDOI      \undefined \def \showDOI       #1{#1}\fi
\ifx \showISBNx    \undefined \def \showISBNx     #1{\unskip}     \fi
\ifx \showISBNxiii \undefined \def \showISBNxiii  #1{\unskip}     \fi
\ifx \showISSN     \undefined \def \showISSN      #1{\unskip}     \fi
\ifx \showLCCN     \undefined \def \showLCCN      #1{\unskip}     \fi
\ifx \shownote     \undefined \def \shownote      #1{#1}          \fi
\ifx \showarticletitle \undefined \def \showarticletitle #1{#1}   \fi
\ifx \showURL      \undefined \def \showURL       {\relax}        \fi
\providecommand\bibfield[2]{#2}
\providecommand\bibinfo[2]{#2}
\providecommand\natexlab[1]{#1}
\providecommand\showeprint[2][]{arXiv:#2}

\bibitem[Alizadeh et~al\mbox{.}(2020)]%
        {alizadeh2020gradient}
\bibfield{author}{\bibinfo{person}{Milad Alizadeh}, \bibinfo{person}{Arash Behboodi}, \bibinfo{person}{Mart van Baalen}, \bibinfo{person}{Christos Louizos}, \bibinfo{person}{Tijmen Blankevoort}, {and} \bibinfo{person}{Max Welling}.} \bibinfo{year}{2020}\natexlab{}.
\newblock \showarticletitle{{G}radient {L}1 regularization for quantization robustness}.
\newblock \bibinfo{journal}{\emph{arXiv preprint arXiv:2002.07520}} (\bibinfo{year}{2020}).
\newblock


\bibitem[Bengio et~al\mbox{.}(2013)]%
        {bengio2013estimating}
\bibfield{author}{\bibinfo{person}{Yoshua Bengio}, \bibinfo{person}{Nicholas L{\'{e}}onard}, {and} \bibinfo{person}{Aaron~C. Courville}.} \bibinfo{year}{2013}\natexlab{}.
\newblock \showarticletitle{{E}stimating or {P}ropagating {G}radients {T}hrough {S}tochastic {N}eurons for {C}onditional {C}omputation}.
\newblock \bibinfo{journal}{\emph{CoRR}}  \bibinfo{volume}{abs/1308.3432} (\bibinfo{year}{2013}).
\newblock
\showeprint[arXiv]{1308.3432}
\urldef\tempurl%
\url{http://arxiv.org/abs/1308.3432}
\showURL{%
\tempurl}


\bibitem[Cai and Vasconcelos(2020)]%
        {cai2020rethinking}
\bibfield{author}{\bibinfo{person}{Zhaowei Cai} {and} \bibinfo{person}{Nuno Vasconcelos}.} \bibinfo{year}{2020}\natexlab{}.
\newblock \showarticletitle{{R}ethinking {D}ifferentiable {S}earch for {M}ixed-Precision {N}eural {N}etworks}. In \bibinfo{booktitle}{\emph{2020 {IEEE/CVF} Conference on Computer Vision and Pattern Recognition, {CVPR} 2020, Seattle, WA, USA, June 13-19, 2020}}. \bibinfo{publisher}{Computer Vision Foundation / {IEEE}}, \bibinfo{pages}{2346--2355}.
\newblock
\urldef\tempurl%
\url{https://doi.org/10.1109/CVPR42600.2020.00242}
\showDOI{\tempurl}


\bibitem[Chen et~al\mbox{.}(2021b)]%
        {chen2021generalized}
\bibfield{author}{\bibinfo{person}{Can Chen}, \bibinfo{person}{Shuhao Zheng}, \bibinfo{person}{Xi Chen}, \bibinfo{person}{Erqun Dong}, \bibinfo{person}{Xue~(Steve) Liu}, \bibinfo{person}{Hao Liu}, {and} \bibinfo{person}{Dejing Dou}.} \bibinfo{year}{2021}\natexlab{b}.
\newblock \showarticletitle{{G}eneralized {D}ata{W}eighting via {C}lass-Level {G}radient {M}anipulation}. In \bibinfo{booktitle}{\emph{Advances in Neural Information Processing Systems 34: Annual Conference on Neural Information Processing Systems 2021, NeurIPS 2021, December 6-14, 2021, virtual}}, \bibfield{editor}{\bibinfo{person}{Marc'Aurelio Ranzato}, \bibinfo{person}{Alina Beygelzimer}, \bibinfo{person}{Yann~N. Dauphin}, \bibinfo{person}{Percy Liang}, {and} \bibinfo{person}{Jennifer~Wortman Vaughan}} (Eds.). \bibinfo{pages}{14097--14109}.
\newblock
\urldef\tempurl%
\url{https://proceedings.neurips.cc/paper/2021/hash/75ebb02f92fc30a8040bbd625af999f1-Abstract.html}
\showURL{%
\tempurl}


\bibitem[Chen et~al\mbox{.}(2021a)]%
        {chen2021towards}
\bibfield{author}{\bibinfo{person}{Weihan Chen}, \bibinfo{person}{Peisong Wang}, {and} \bibinfo{person}{Jian Cheng}.} \bibinfo{year}{2021}\natexlab{a}.
\newblock \showarticletitle{{T}owards {M}ixed-Precision {Q}uantization of {N}eural {N}etworks via {C}onstrained {O}ptimization}. In \bibinfo{booktitle}{\emph{2021 {IEEE/CVF} International Conference on Computer Vision, {ICCV} 2021, Montreal, QC, Canada, October 10-17, 2021}}. \bibinfo{publisher}{{IEEE}}, \bibinfo{pages}{5330--5339}.
\newblock
\urldef\tempurl%
\url{https://doi.org/10.1109/ICCV48922.2021.00530}
\showDOI{\tempurl}


\bibitem[Choi et~al\mbox{.}(2018)]%
        {choi2018pact}
\bibfield{author}{\bibinfo{person}{Jungwook Choi}, \bibinfo{person}{Zhuo Wang}, \bibinfo{person}{Swagath Venkataramani}, \bibinfo{person}{Pierce~I{-}Jen Chuang}, \bibinfo{person}{Vijayalakshmi Srinivasan}, {and} \bibinfo{person}{Kailash Gopalakrishnan}.} \bibinfo{year}{2018}\natexlab{}.
\newblock \showarticletitle{{PACT:} {P}arameterized {C}lipping {A}ctivation for {Q}uantized {N}eural {N}etworks}.
\newblock \bibinfo{journal}{\emph{CoRR}}  \bibinfo{volume}{abs/1805.06085} (\bibinfo{year}{2018}).
\newblock
\showeprint[arXiv]{1805.06085}
\urldef\tempurl%
\url{http://arxiv.org/abs/1805.06085}
\showURL{%
\tempurl}


\bibitem[Deng et~al\mbox{.}(2009)]%
        {deng2009imagenet}
\bibfield{author}{\bibinfo{person}{Jia Deng}, \bibinfo{person}{Wei Dong}, \bibinfo{person}{Richard Socher}, \bibinfo{person}{Li{-}Jia Li}, \bibinfo{person}{Kai Li}, {and} \bibinfo{person}{Li Fei{-}Fei}.} \bibinfo{year}{2009}\natexlab{}.
\newblock \showarticletitle{{I}mage{N}et: {A} large-scale hierarchical image database}. In \bibinfo{booktitle}{\emph{2009 {IEEE} Computer Society Conference on Computer Vision and Pattern Recognition {(CVPR} 2009), 20-25 June 2009, Miami, Florida, {USA}}}. \bibinfo{publisher}{{IEEE} Computer Society}, \bibinfo{pages}{248--255}.
\newblock
\urldef\tempurl%
\url{https://doi.org/10.1109/CVPR.2009.5206848}
\showDOI{\tempurl}


\bibitem[Dong et~al\mbox{.}(2020)]%
        {dong2020hawq}
\bibfield{author}{\bibinfo{person}{Zhen Dong}, \bibinfo{person}{Zhewei Yao}, \bibinfo{person}{Daiyaan Arfeen}, \bibinfo{person}{Amir Gholami}, \bibinfo{person}{Michael~W. Mahoney}, {and} \bibinfo{person}{Kurt Keutzer}.} \bibinfo{year}{2020}\natexlab{}.
\newblock \showarticletitle{{HAWQ-V2:} {H}essian {A}ware trace-Weighted {Q}uantization of {N}eural {N}etworks}. In \bibinfo{booktitle}{\emph{Advances in Neural Information Processing Systems 33: Annual Conference on Neural Information Processing Systems 2020, NeurIPS 2020, December 6-12, 2020, virtual}}, \bibfield{editor}{\bibinfo{person}{Hugo Larochelle}, \bibinfo{person}{Marc'Aurelio Ranzato}, \bibinfo{person}{Raia Hadsell}, \bibinfo{person}{Maria{-}Florina Balcan}, {and} \bibinfo{person}{Hsuan{-}Tien Lin}} (Eds.).
\newblock
\urldef\tempurl%
\url{https://proceedings.neurips.cc/paper/2020/hash/d77c703536718b95308130ff2e5cf9ee-Abstract.html}
\showURL{%
\tempurl}


\bibitem[Dong et~al\mbox{.}(2019)]%
        {dong2019hawq}
\bibfield{author}{\bibinfo{person}{Zhen Dong}, \bibinfo{person}{Zhewei Yao}, \bibinfo{person}{Amir Gholami}, \bibinfo{person}{Michael~W. Mahoney}, {and} \bibinfo{person}{Kurt Keutzer}.} \bibinfo{year}{2019}\natexlab{}.
\newblock \showarticletitle{{HAWQ:} {H}essian {A}{W}are {Q}uantization of {N}eural {N}etworks {W}ith {M}ixed-Precision}. In \bibinfo{booktitle}{\emph{2019 {IEEE/CVF} International Conference on Computer Vision, {ICCV} 2019, Seoul, Korea (South), October 27 - November 2, 2019}}. \bibinfo{publisher}{{IEEE}}, \bibinfo{pages}{293--302}.
\newblock
\urldef\tempurl%
\url{https://doi.org/10.1109/ICCV.2019.00038}
\showDOI{\tempurl}


\bibitem[Dudoit et~al\mbox{.}(2002)]%
        {dudoit2002comparison}
\bibfield{author}{\bibinfo{person}{Sandrine Dudoit}, \bibinfo{person}{Jane Fridlyand}, {and} \bibinfo{person}{Terence~P Speed}.} \bibinfo{year}{2002}\natexlab{}.
\newblock \showarticletitle{Comparison of discrimination methods for the classification of tumors using gene expression data}.
\newblock \bibinfo{journal}{\emph{Journal of the American statistical association}} \bibinfo{volume}{97}, \bibinfo{number}{457} (\bibinfo{year}{2002}), \bibinfo{pages}{77--87}.
\newblock


\bibitem[Elthakeb et~al\mbox{.}(2020)]%
        {elthakeb2020releq}
\bibfield{author}{\bibinfo{person}{Ahmed~T. Elthakeb}, \bibinfo{person}{Prannoy Pilligundla}, \bibinfo{person}{Fatemehsadat Mireshghallah}, \bibinfo{person}{Amir Yazdanbakhsh}, {and} \bibinfo{person}{Hadi Esmaeilzadeh}.} \bibinfo{year}{2020}\natexlab{}.
\newblock \showarticletitle{{R}e{L}e{Q} : {A} {R}einforcement {L}earning {A}pproach for {A}utomatic {D}eep {Q}uantization of {N}eural {N}etworks}.
\newblock \bibinfo{journal}{\emph{{IEEE} Micro}} \bibinfo{volume}{40}, \bibinfo{number}{5} (\bibinfo{year}{2020}), \bibinfo{pages}{37--45}.
\newblock
\urldef\tempurl%
\url{https://doi.org/10.1109/MM.2020.3009475}
\showDOI{\tempurl}


\bibitem[Esser et~al\mbox{.}(2020)]%
        {esser2020learned}
\bibfield{author}{\bibinfo{person}{Steven~K. Esser}, \bibinfo{person}{Jeffrey~L. McKinstry}, \bibinfo{person}{Deepika Bablani}, \bibinfo{person}{Rathinakumar Appuswamy}, {and} \bibinfo{person}{Dharmendra~S. Modha}.} \bibinfo{year}{2020}\natexlab{}.
\newblock \showarticletitle{{L}earned {S}tep {S}ize quantization}. In \bibinfo{booktitle}{\emph{8th International Conference on Learning Representations, {ICLR} 2020, Addis Ababa, Ethiopia, April 26-30, 2020}}. \bibinfo{publisher}{OpenReview.net}.
\newblock
\urldef\tempurl%
\url{https://openreview.net/forum?id=rkgO66VKDS}
\showURL{%
\tempurl}


\bibitem[Gong et~al\mbox{.}(2019)]%
        {gong2019differentiable}
\bibfield{author}{\bibinfo{person}{Ruihao Gong}, \bibinfo{person}{Xianglong Liu}, \bibinfo{person}{Shenghu Jiang}, \bibinfo{person}{Tianxiang Li}, \bibinfo{person}{Peng Hu}, \bibinfo{person}{Jiazhen Lin}, \bibinfo{person}{Fengwei Yu}, {and} \bibinfo{person}{Junjie Yan}.} \bibinfo{year}{2019}\natexlab{}.
\newblock \showarticletitle{{D}ifferentiable {S}oft {Q}uantization: {B}ridging {F}ull-Precision and {L}ow-Bit {N}eural {N}etworks}. In \bibinfo{booktitle}{\emph{2019 {IEEE/CVF} International Conference on Computer Vision, {ICCV} 2019, Seoul, Korea (South), October 27 - November 2, 2019}}. \bibinfo{publisher}{{IEEE}}, \bibinfo{pages}{4851--4860}.
\newblock
\urldef\tempurl%
\url{https://doi.org/10.1109/ICCV.2019.00495}
\showDOI{\tempurl}


\bibitem[Guo et~al\mbox{.}(2020)]%
        {guo2020single}
\bibfield{author}{\bibinfo{person}{Zichao Guo}, \bibinfo{person}{Xiangyu Zhang}, \bibinfo{person}{Haoyuan Mu}, \bibinfo{person}{Wen Heng}, \bibinfo{person}{Zechun Liu}, \bibinfo{person}{Yichen Wei}, {and} \bibinfo{person}{Jian Sun}.} \bibinfo{year}{2020}\natexlab{}.
\newblock \showarticletitle{{S}ingle {P}ath {O}ne-Shot {N}eural {A}rchitecture {S}earch with {U}niform {S}ampling}. In \bibinfo{booktitle}{\emph{Computer Vision - {ECCV} 2020 - 16th European Conference, Glasgow, UK, August 23-28, 2020, Proceedings, Part {XVI}}} \emph{(\bibinfo{series}{Lecture Notes in Computer Science}, Vol.~\bibinfo{volume}{12361})}, \bibfield{editor}{\bibinfo{person}{Andrea Vedaldi}, \bibinfo{person}{Horst Bischof}, \bibinfo{person}{Thomas Brox}, {and} \bibinfo{person}{Jan{-}Michael Frahm}} (Eds.). \bibinfo{publisher}{Springer}, \bibinfo{pages}{544--560}.
\newblock
\urldef\tempurl%
\url{https://doi.org/10.1007/978-3-030-58517-4\_32}
\showDOI{\tempurl}


\bibitem[Habi et~al\mbox{.}(2020)]%
        {habi2020hmq}
\bibfield{author}{\bibinfo{person}{Hai~Victor Habi}, \bibinfo{person}{Roy~H. Jennings}, {and} \bibinfo{person}{Arnon Netzer}.} \bibinfo{year}{2020}\natexlab{}.
\newblock \showarticletitle{{HMQ:} {H}ardware {F}riendly {M}ixed {P}recision {Q}uantization {B}lock for {C}{N}{N}s}. In \bibinfo{booktitle}{\emph{Computer Vision - {ECCV} 2020 - 16th European Conference, Glasgow, UK, August 23-28, 2020, Proceedings, Part {XXVI}}} \emph{(\bibinfo{series}{Lecture Notes in Computer Science}, Vol.~\bibinfo{volume}{12371})}, \bibfield{editor}{\bibinfo{person}{Andrea Vedaldi}, \bibinfo{person}{Horst Bischof}, \bibinfo{person}{Thomas Brox}, {and} \bibinfo{person}{Jan{-}Michael Frahm}} (Eds.). \bibinfo{publisher}{Springer}, \bibinfo{pages}{448--463}.
\newblock
\urldef\tempurl%
\url{https://doi.org/10.1007/978-3-030-58574-7\_27}
\showDOI{\tempurl}


\bibitem[Hadsell et~al\mbox{.}(2006)]%
        {hadsell2006dimensionality}
\bibfield{author}{\bibinfo{person}{Raia Hadsell}, \bibinfo{person}{Sumit Chopra}, {and} \bibinfo{person}{Yann LeCun}.} \bibinfo{year}{2006}\natexlab{}.
\newblock \showarticletitle{{D}imensionality {R}eduction by {L}earning an {I}nvariant {M}apping}. In \bibinfo{booktitle}{\emph{2006 {IEEE} Computer Society Conference on Computer Vision and Pattern Recognition {(CVPR} 2006), 17-22 June 2006, New York, NY, {USA}}}. \bibinfo{publisher}{{IEEE} Computer Society}, \bibinfo{pages}{1735--1742}.
\newblock
\urldef\tempurl%
\url{https://doi.org/10.1109/CVPR.2006.100}
\showDOI{\tempurl}


\bibitem[Han et~al\mbox{.}(2021)]%
        {han2021improving}
\bibfield{author}{\bibinfo{person}{Tiantian Han}, \bibinfo{person}{Dong Li}, \bibinfo{person}{Ji Liu}, \bibinfo{person}{Lu Tian}, {and} \bibinfo{person}{Yi Shan}.} \bibinfo{year}{2021}\natexlab{}.
\newblock \showarticletitle{{I}mproving {L}ow-Precision {N}etwork {Q}uantization via {B}in {R}egularization}. In \bibinfo{booktitle}{\emph{2021 {IEEE/CVF} International Conference on Computer Vision, {ICCV} 2021, Montreal, QC, Canada, October 10-17, 2021}}. \bibinfo{publisher}{{IEEE}}, \bibinfo{pages}{5241--5250}.
\newblock
\urldef\tempurl%
\url{https://doi.org/10.1109/ICCV48922.2021.00521}
\showDOI{\tempurl}


\bibitem[He et~al\mbox{.}(2016)]%
        {he2016deep}
\bibfield{author}{\bibinfo{person}{Kaiming He}, \bibinfo{person}{Xiangyu Zhang}, \bibinfo{person}{Shaoqing Ren}, {and} \bibinfo{person}{Jian Sun}.} \bibinfo{year}{2016}\natexlab{}.
\newblock \showarticletitle{{D}eep {R}esidual {L}earning for {I}mage {R}ecognition}. In \bibinfo{booktitle}{\emph{2016 {IEEE} Conference on Computer Vision and Pattern Recognition, {CVPR} 2016, Las Vegas, NV, USA, June 27-30, 2016}}. \bibinfo{publisher}{{IEEE} Computer Society}, \bibinfo{pages}{770--778}.
\newblock
\urldef\tempurl%
\url{https://doi.org/10.1109/CVPR.2016.90}
\showDOI{\tempurl}


\bibitem[Hinton et~al\mbox{.}(2015)]%
        {hinton2015distilling}
\bibfield{author}{\bibinfo{person}{Geoffrey~E. Hinton}, \bibinfo{person}{Oriol Vinyals}, {and} \bibinfo{person}{Jeffrey Dean}.} \bibinfo{year}{2015}\natexlab{}.
\newblock \showarticletitle{{D}istilling the {K}nowledge in a {N}eural {N}etwork}.
\newblock \bibinfo{journal}{\emph{CoRR}}  \bibinfo{volume}{abs/1503.02531} (\bibinfo{year}{2015}).
\newblock
\showeprint[arXiv]{1503.02531}
\urldef\tempurl%
\url{http://arxiv.org/abs/1503.02531}
\showURL{%
\tempurl}


\bibitem[Howard et~al\mbox{.}(2017)]%
        {howard2017mobilenets}
\bibfield{author}{\bibinfo{person}{Andrew~G. Howard}, \bibinfo{person}{Menglong Zhu}, \bibinfo{person}{Bo Chen}, \bibinfo{person}{Dmitry Kalenichenko}, \bibinfo{person}{Weijun Wang}, \bibinfo{person}{Tobias Weyand}, \bibinfo{person}{Marco Andreetto}, {and} \bibinfo{person}{Hartwig Adam}.} \bibinfo{year}{2017}\natexlab{}.
\newblock \showarticletitle{{M}obile{N}ets: {E}fficient {C}onvolutional {N}eural {N}etworks for {M}obile {V}ision {A}pplications}.
\newblock \bibinfo{journal}{\emph{CoRR}}  \bibinfo{volume}{abs/1704.04861} (\bibinfo{year}{2017}).
\newblock
\showeprint[arXiv]{1704.04861}
\urldef\tempurl%
\url{http://arxiv.org/abs/1704.04861}
\showURL{%
\tempurl}


\bibitem[Huang et~al\mbox{.}(2022)]%
        {huang2022sdq}
\bibfield{author}{\bibinfo{person}{Xijie Huang}, \bibinfo{person}{Zhiqiang Shen}, \bibinfo{person}{Shichao Li}, \bibinfo{person}{Zechun Liu}, \bibinfo{person}{Xianghong Hu}, \bibinfo{person}{Jeffry Wicaksana}, \bibinfo{person}{Eric~P. Xing}, {and} \bibinfo{person}{Kwang{-}Ting Cheng}.} \bibinfo{year}{2022}\natexlab{}.
\newblock \showarticletitle{{SDQ:} {S}tochastic {D}ifferentiable {Q}uantization with {M}ixed {P}recision}. In \bibinfo{booktitle}{\emph{International Conference on Machine Learning, {ICML} 2022, 17-23 July 2022, Baltimore, Maryland, {USA}}} \emph{(\bibinfo{series}{Proceedings of Machine Learning Research}, Vol.~\bibinfo{volume}{162})}, \bibfield{editor}{\bibinfo{person}{Kamalika Chaudhuri}, \bibinfo{person}{Stefanie Jegelka}, \bibinfo{person}{Le~Song}, \bibinfo{person}{Csaba Szepesv{\'{a}}ri}, \bibinfo{person}{Gang Niu}, {and} \bibinfo{person}{Sivan Sabato}} (Eds.). \bibinfo{publisher}{{PMLR}}, \bibinfo{pages}{9295--9309}.
\newblock
\urldef\tempurl%
\url{https://proceedings.mlr.press/v162/huang22h.html}
\showURL{%
\tempurl}


\bibitem[Hubara et~al\mbox{.}(2021)]%
        {hubara2021accurate}
\bibfield{author}{\bibinfo{person}{Itay Hubara}, \bibinfo{person}{Yury Nahshan}, \bibinfo{person}{Yair Hanani}, \bibinfo{person}{Ron Banner}, {and} \bibinfo{person}{Daniel Soudry}.} \bibinfo{year}{2021}\natexlab{}.
\newblock \showarticletitle{Accurate post training quantization with small calibration sets}. In \bibinfo{booktitle}{\emph{International Conference on Machine Learning}}. PMLR, \bibinfo{pages}{4466--4475}.
\newblock


\bibitem[Krause et~al\mbox{.}(2013)]%
        {KrauseStarkDengFei-Fei_3DRR2013}
\bibfield{author}{\bibinfo{person}{Jonathan Krause}, \bibinfo{person}{Michael Stark}, \bibinfo{person}{Jia Deng}, {and} \bibinfo{person}{Li Fei{-}Fei}.} \bibinfo{year}{2013}\natexlab{}.
\newblock \showarticletitle{3{D} {O}bject {R}epresentations for {F}ine-Grained {C}ategorization}. In \bibinfo{booktitle}{\emph{2013 {IEEE} International Conference on Computer Vision Workshops, {ICCV} Workshops 2013, Sydney, Australia, December 1-8, 2013}}. \bibinfo{publisher}{{IEEE} Computer Society}, \bibinfo{pages}{554--561}.
\newblock
\urldef\tempurl%
\url{https://doi.org/10.1109/ICCVW.2013.77}
\showDOI{\tempurl}


\bibitem[Krizhevsky et~al\mbox{.}(2009)]%
        {krizhevsky2009learning}
\bibfield{author}{\bibinfo{person}{Alex Krizhevsky}, \bibinfo{person}{Geoffrey Hinton}, {et~al\mbox{.}}} \bibinfo{year}{2009}\natexlab{}.
\newblock \showarticletitle{{L}earning multiple layers of features from tiny images}.
\newblock  (\bibinfo{year}{2009}).
\newblock


\bibitem[Li et~al\mbox{.}(2023)]%
        {li2023hard}
\bibfield{author}{\bibinfo{person}{Huantong Li}, \bibinfo{person}{Xiangmiao Wu}, \bibinfo{person}{Fanbing Lv}, \bibinfo{person}{Daihai Liao}, \bibinfo{person}{Thomas~H Li}, \bibinfo{person}{Yonggang Zhang}, \bibinfo{person}{Bo Han}, {and} \bibinfo{person}{Mingkui Tan}.} \bibinfo{year}{2023}\natexlab{}.
\newblock \showarticletitle{Hard Sample Matters a Lot in Zero-Shot Quantization}. In \bibinfo{booktitle}{\emph{Proceedings of the IEEE/CVF Conference on Computer Vision and Pattern Recognition}}. \bibinfo{pages}{24417--24426}.
\newblock


\bibitem[Liu et~al\mbox{.}(2023)]%
        {liu2023single}
\bibfield{author}{\bibinfo{person}{Jing Liu}, \bibinfo{person}{Bohan Zhuang}, \bibinfo{person}{Peng Chen}, \bibinfo{person}{Chunhua Shen}, \bibinfo{person}{Jianfei Cai}, {and} \bibinfo{person}{Mingkui Tan}.} \bibinfo{year}{2023}\natexlab{}.
\newblock \showarticletitle{Single-path bit sharing for automatic loss-aware model compression}.
\newblock \bibinfo{journal}{\emph{IEEE Transactions on Pattern Analysis and Machine Intelligence}} (\bibinfo{year}{2023}).
\newblock


\bibitem[Liu et~al\mbox{.}(2016)]%
        {liu2016large}
\bibfield{author}{\bibinfo{person}{Weiyang Liu}, \bibinfo{person}{Yandong Wen}, \bibinfo{person}{Zhiding Yu}, {and} \bibinfo{person}{Meng Yang}.} \bibinfo{year}{2016}\natexlab{}.
\newblock \showarticletitle{{L}arge-Margin {S}oftmax {L}oss for {C}onvolutional {N}eural {N}etworks}. In \bibinfo{booktitle}{\emph{Proceedings of the 33nd International Conference on Machine Learning, {ICML} 2016, New York City, NY, USA, June 19-24, 2016}} \emph{(\bibinfo{series}{{JMLR} Workshop and Conference Proceedings}, Vol.~\bibinfo{volume}{48})}, \bibfield{editor}{\bibinfo{person}{Maria{-}Florina Balcan} {and} \bibinfo{person}{Kilian~Q. Weinberger}} (Eds.). \bibinfo{publisher}{JMLR.org}, \bibinfo{pages}{507--516}.
\newblock
\urldef\tempurl%
\url{http://proceedings.mlr.press/v48/liud16.html}
\showURL{%
\tempurl}


\bibitem[Liu et~al\mbox{.}(2022)]%
        {liu2022nonuniform}
\bibfield{author}{\bibinfo{person}{Zechun Liu}, \bibinfo{person}{Kwang-Ting Cheng}, \bibinfo{person}{Dong Huang}, \bibinfo{person}{Eric~P Xing}, {and} \bibinfo{person}{Zhiqiang Shen}.} \bibinfo{year}{2022}\natexlab{}.
\newblock \showarticletitle{Nonuniform-to-uniform quantization: Towards accurate quantization via generalized straight-through estimation}. In \bibinfo{booktitle}{\emph{Proceedings of the IEEE/CVF Conference on Computer Vision and Pattern Recognition}}. \bibinfo{pages}{4942--4952}.
\newblock


\bibitem[Liu et~al\mbox{.}(2019)]%
        {liu2018rethinking}
\bibfield{author}{\bibinfo{person}{Zhuang Liu}, \bibinfo{person}{Mingjie Sun}, \bibinfo{person}{Tinghui Zhou}, \bibinfo{person}{Gao Huang}, {and} \bibinfo{person}{Trevor Darrell}.} \bibinfo{year}{2019}\natexlab{}.
\newblock \showarticletitle{{R}ethinking the {V}alue of {N}etwork {P}runing}. In \bibinfo{booktitle}{\emph{7th International Conference on Learning Representations, {ICLR} 2019, New Orleans, LA, USA, May 6-9, 2019}}. \bibinfo{publisher}{OpenReview.net}.
\newblock
\urldef\tempurl%
\url{https://openreview.net/forum?id=rJlnB3C5Ym}
\showURL{%
\tempurl}


\bibitem[Nagel et~al\mbox{.}(2020)]%
        {nagel2020up}
\bibfield{author}{\bibinfo{person}{Markus Nagel}, \bibinfo{person}{Rana~Ali Amjad}, \bibinfo{person}{Mart Van~Baalen}, \bibinfo{person}{Christos Louizos}, {and} \bibinfo{person}{Tijmen Blankevoort}.} \bibinfo{year}{2020}\natexlab{}.
\newblock \showarticletitle{Up or down? adaptive rounding for post-training quantization}. In \bibinfo{booktitle}{\emph{International Conference on Machine Learning}}. PMLR, \bibinfo{pages}{7197--7206}.
\newblock


\bibitem[Park and Yoo(2020)]%
        {park2020profit}
\bibfield{author}{\bibinfo{person}{Eunhyeok Park} {and} \bibinfo{person}{Sungjoo Yoo}.} \bibinfo{year}{2020}\natexlab{}.
\newblock \showarticletitle{{PROFIT:} {A} {N}ovel {T}raining {M}ethod for sub-4-bit {M}obile{N}et {M}odels}. In \bibinfo{booktitle}{\emph{Computer Vision - {ECCV} 2020 - 16th European Conference, Glasgow, UK, August 23-28, 2020, Proceedings, Part {VI}}} \emph{(\bibinfo{series}{Lecture Notes in Computer Science}, Vol.~\bibinfo{volume}{12351})}, \bibfield{editor}{\bibinfo{person}{Andrea Vedaldi}, \bibinfo{person}{Horst Bischof}, \bibinfo{person}{Thomas Brox}, {and} \bibinfo{person}{Jan{-}Michael Frahm}} (Eds.). \bibinfo{publisher}{Springer}, \bibinfo{pages}{430--446}.
\newblock
\urldef\tempurl%
\url{https://doi.org/10.1007/978-3-030-58539-6\_26}
\showDOI{\tempurl}


\bibitem[Ranasinghe et~al\mbox{.}(2021)]%
        {ranasinghe2021orthogonal}
\bibfield{author}{\bibinfo{person}{Kanchana Ranasinghe}, \bibinfo{person}{Muzammal Naseer}, \bibinfo{person}{Munawar Hayat}, \bibinfo{person}{Salman~H. Khan}, {and} \bibinfo{person}{Fahad~Shahbaz Khan}.} \bibinfo{year}{2021}\natexlab{}.
\newblock \showarticletitle{{O}rthogonal {P}rojection {L}oss}. In \bibinfo{booktitle}{\emph{2021 {IEEE/CVF} International Conference on Computer Vision, {ICCV} 2021, Montreal, QC, Canada, October 10-17, 2021}}. \bibinfo{publisher}{{IEEE}}, \bibinfo{pages}{12313--12323}.
\newblock
\urldef\tempurl%
\url{https://doi.org/10.1109/ICCV48922.2021.01211}
\showDOI{\tempurl}


\bibitem[Redmon et~al\mbox{.}(2016)]%
        {redmon2016you}
\bibfield{author}{\bibinfo{person}{Joseph Redmon}, \bibinfo{person}{Santosh Divvala}, \bibinfo{person}{Ross Girshick}, {and} \bibinfo{person}{Ali Farhadi}.} \bibinfo{year}{2016}\natexlab{}.
\newblock \showarticletitle{You only look once: Unified, real-time object detection}. In \bibinfo{booktitle}{\emph{Proceedings of the IEEE conference on computer vision and pattern recognition}}. \bibinfo{pages}{779--788}.
\newblock


\bibitem[Ren et~al\mbox{.}(2015)]%
        {ren2015faster}
\bibfield{author}{\bibinfo{person}{Shaoqing Ren}, \bibinfo{person}{Kaiming He}, \bibinfo{person}{Ross Girshick}, {and} \bibinfo{person}{Jian Sun}.} \bibinfo{year}{2015}\natexlab{}.
\newblock \showarticletitle{Faster r-cnn: Towards real-time object detection with region proposal networks}.
\newblock \bibinfo{journal}{\emph{Advances in neural information processing systems}}  \bibinfo{volume}{28} (\bibinfo{year}{2015}).
\newblock


\bibitem[Sandler et~al\mbox{.}(2018)]%
        {sandler2018mobilenetv2}
\bibfield{author}{\bibinfo{person}{Mark Sandler}, \bibinfo{person}{Andrew~G. Howard}, \bibinfo{person}{Menglong Zhu}, \bibinfo{person}{Andrey Zhmoginov}, {and} \bibinfo{person}{Liang{-}Chieh Chen}.} \bibinfo{year}{2018}\natexlab{}.
\newblock \showarticletitle{{M}obile{N}et{V}2: {I}nverted {R}esiduals and {L}inear {B}ottlenecks}. In \bibinfo{booktitle}{\emph{2018 {IEEE} Conference on Computer Vision and Pattern Recognition, {CVPR} 2018, Salt Lake City, UT, USA, June 18-22, 2018}}. \bibinfo{publisher}{Computer Vision Foundation / {IEEE} Computer Society}, \bibinfo{pages}{4510--4520}.
\newblock
\urldef\tempurl%
\url{https://doi.org/10.1109/CVPR.2018.00474}
\showDOI{\tempurl}


\bibitem[Selvaraju et~al\mbox{.}(2020)]%
        {selvaraju2017grad}
\bibfield{author}{\bibinfo{person}{Ramprasaath~R. Selvaraju}, \bibinfo{person}{Michael Cogswell}, \bibinfo{person}{Abhishek Das}, \bibinfo{person}{Ramakrishna Vedantam}, \bibinfo{person}{Devi Parikh}, {and} \bibinfo{person}{Dhruv Batra}.} \bibinfo{year}{2020}\natexlab{}.
\newblock \showarticletitle{{G}rad-{C}{A}{M}: {V}isual {E}xplanations from {D}eep {N}etworks via {G}radient-Based {L}ocalization}.
\newblock \bibinfo{journal}{\emph{Int. J. Comput. Vis.}} \bibinfo{volume}{128}, \bibinfo{number}{2} (\bibinfo{year}{2020}), \bibinfo{pages}{336--359}.
\newblock
\urldef\tempurl%
\url{https://doi.org/10.1007/s11263-019-01228-7}
\showDOI{\tempurl}


\bibitem[Shi et~al\mbox{.}(2019)]%
        {shi2019watch}
\bibfield{author}{\bibinfo{person}{Xiangxi Shi}, \bibinfo{person}{Jianfei Cai}, \bibinfo{person}{Shafiq Joty}, {and} \bibinfo{person}{Jiuxiang Gu}.} \bibinfo{year}{2019}\natexlab{}.
\newblock \showarticletitle{Watch it twice: Video captioning with a refocused video encoder}. In \bibinfo{booktitle}{\emph{Proceedings of the 27th ACM International Conference on Multimedia}}. \bibinfo{pages}{818--826}.
\newblock


\bibitem[Shkolnik et~al\mbox{.}(2020)]%
        {chmiel2020robust}
\bibfield{author}{\bibinfo{person}{Moran Shkolnik}, \bibinfo{person}{Brian Chmiel}, \bibinfo{person}{Ron Banner}, \bibinfo{person}{Gil Shomron}, \bibinfo{person}{Yury Nahshan}, \bibinfo{person}{Alex~M. Bronstein}, {and} \bibinfo{person}{Uri~C. Weiser}.} \bibinfo{year}{2020}\natexlab{}.
\newblock \showarticletitle{{R}obust {Q}uantization: {O}ne {M}odel to {R}ule {T}hem {A}ll}. In \bibinfo{booktitle}{\emph{Advances in Neural Information Processing Systems 33: Annual Conference on Neural Information Processing Systems 2020, NeurIPS 2020, December 6-12, 2020, virtual}}, \bibfield{editor}{\bibinfo{person}{Hugo Larochelle}, \bibinfo{person}{Marc'Aurelio Ranzato}, \bibinfo{person}{Raia Hadsell}, \bibinfo{person}{Maria{-}Florina Balcan}, {and} \bibinfo{person}{Hsuan{-}Tien Lin}} (Eds.).
\newblock
\urldef\tempurl%
\url{https://proceedings.neurips.cc/paper/2020/hash/3948ead63a9f2944218de038d8934305-Abstract.html}
\showURL{%
\tempurl}


\bibitem[Sutskever et~al\mbox{.}(2013)]%
        {sutskever2013importance}
\bibfield{author}{\bibinfo{person}{Ilya Sutskever}, \bibinfo{person}{James Martens}, \bibinfo{person}{George~E. Dahl}, {and} \bibinfo{person}{Geoffrey~E. Hinton}.} \bibinfo{year}{2013}\natexlab{}.
\newblock \showarticletitle{{O}n the importance of initialization and momentum in deep learning}. In \bibinfo{booktitle}{\emph{Proceedings of the 30th International Conference on Machine Learning, {ICML} 2013, Atlanta, GA, USA, 16-21 June 2013}} \emph{(\bibinfo{series}{{JMLR} Workshop and Conference Proceedings}, Vol.~\bibinfo{volume}{28})}. \bibinfo{publisher}{JMLR.org}, \bibinfo{pages}{1139--1147}.
\newblock
\urldef\tempurl%
\url{http://proceedings.mlr.press/v28/sutskever13.html}
\showURL{%
\tempurl}


\bibitem[Tang et~al\mbox{.}(2022a)]%
        {tang2022mixed}
\bibfield{author}{\bibinfo{person}{Chen Tang}, \bibinfo{person}{Kai Ouyang}, \bibinfo{person}{Zhi Wang}, \bibinfo{person}{Yifei Zhu}, \bibinfo{person}{Wen Ji}, \bibinfo{person}{Yaowei Wang}, {and} \bibinfo{person}{Wenwu Zhu}.} \bibinfo{year}{2022}\natexlab{a}.
\newblock \showarticletitle{{M}ixed-Precision {N}eural {N}etwork {Q}uantization via {L}earned {L}ayer-Wise {I}mportance}. In \bibinfo{booktitle}{\emph{Computer Vision - {ECCV} 2022 - 17th European Conference, Tel Aviv, Israel, October 23-27, 2022, Proceedings, Part {XI}}} \emph{(\bibinfo{series}{Lecture Notes in Computer Science}, Vol.~\bibinfo{volume}{13671})}, \bibfield{editor}{\bibinfo{person}{Shai Avidan}, \bibinfo{person}{Gabriel~J. Brostow}, \bibinfo{person}{Moustapha Ciss{\'{e}}}, \bibinfo{person}{Giovanni~Maria Farinella}, {and} \bibinfo{person}{Tal Hassner}} (Eds.). \bibinfo{publisher}{Springer}, \bibinfo{pages}{259--275}.
\newblock
\urldef\tempurl%
\url{https://doi.org/10.1007/978-3-031-20083-0\_16}
\showDOI{\tempurl}


\bibitem[Tang et~al\mbox{.}(2022b)]%
        {tang2022arbitrary}
\bibfield{author}{\bibinfo{person}{Chen Tang}, \bibinfo{person}{Haoyu Zhai}, \bibinfo{person}{Kai Ouyang}, \bibinfo{person}{Zhi Wang}, \bibinfo{person}{Yifei Zhu}, {and} \bibinfo{person}{Wenwu Zhu}.} \bibinfo{year}{2022}\natexlab{b}.
\newblock \showarticletitle{Arbitrary Bit-width Network: A Joint Layer-Wise Quantization and Adaptive Inference Approach}. In \bibinfo{booktitle}{\emph{Proceedings of the 30th ACM International Conference on Multimedia}}. \bibinfo{pages}{2899--2908}.
\newblock


\bibitem[Van~der Maaten and Hinton(2008)]%
        {van2008visualizing}
\bibfield{author}{\bibinfo{person}{Laurens Van~der Maaten} {and} \bibinfo{person}{Geoffrey Hinton}.} \bibinfo{year}{2008}\natexlab{}.
\newblock \showarticletitle{{V}isualizing data using t-{S}{N}{E}.}
\newblock \bibinfo{journal}{\emph{Journal of machine learning research}} \bibinfo{volume}{9}, \bibinfo{number}{11} (\bibinfo{year}{2008}).
\newblock


\bibitem[Veit et~al\mbox{.}(2016)]%
        {veit2016residual}
\bibfield{author}{\bibinfo{person}{Andreas Veit}, \bibinfo{person}{Michael~J. Wilber}, {and} \bibinfo{person}{Serge~J. Belongie}.} \bibinfo{year}{2016}\natexlab{}.
\newblock \showarticletitle{{R}esidual {N}etworks {B}ehave {L}ike {E}nsembles of {R}elatively {S}hallow {N}etworks}. In \bibinfo{booktitle}{\emph{Advances in Neural Information Processing Systems 29: Annual Conference on Neural Information Processing Systems 2016, December 5-10, 2016, Barcelona, Spain}}, \bibfield{editor}{\bibinfo{person}{Daniel~D. Lee}, \bibinfo{person}{Masashi Sugiyama}, \bibinfo{person}{Ulrike von Luxburg}, \bibinfo{person}{Isabelle Guyon}, {and} \bibinfo{person}{Roman Garnett}} (Eds.). \bibinfo{pages}{550--558}.
\newblock
\urldef\tempurl%
\url{https://proceedings.neurips.cc/paper/2016/hash/37bc2f75bf1bcfe8450a1a41c200364c-Abstract.html}
\showURL{%
\tempurl}


\bibitem[Wan et~al\mbox{.}(2022)]%
        {wan2022shaping}
\bibfield{author}{\bibinfo{person}{Weitao Wan}, \bibinfo{person}{Cheng Yu}, \bibinfo{person}{Jiansheng Chen}, \bibinfo{person}{Tong Wu}, \bibinfo{person}{Yuanyi Zhong}, {and} \bibinfo{person}{Ming-Hsuan Yang}.} \bibinfo{year}{2022}\natexlab{}.
\newblock \showarticletitle{Shaping deep feature space towards gaussian mixture for visual classification}.
\newblock \bibinfo{journal}{\emph{IEEE transactions on pattern analysis and machine intelligence}} \bibinfo{volume}{45}, \bibinfo{number}{2} (\bibinfo{year}{2022}), \bibinfo{pages}{2430--2444}.
\newblock


\bibitem[Wan et~al\mbox{.}(2018)]%
        {wan2018rethinking}
\bibfield{author}{\bibinfo{person}{Weitao Wan}, \bibinfo{person}{Yuanyi Zhong}, \bibinfo{person}{Tianpeng Li}, {and} \bibinfo{person}{Jiansheng Chen}.} \bibinfo{year}{2018}\natexlab{}.
\newblock \showarticletitle{{R}ethinking {F}eature {D}istribution for {L}oss {F}unctions in {I}mage {C}lassification}. In \bibinfo{booktitle}{\emph{2018 {IEEE} Conference on Computer Vision and Pattern Recognition, {CVPR} 2018, Salt Lake City, UT, USA, June 18-22, 2018}}. \bibinfo{publisher}{Computer Vision Foundation / {IEEE} Computer Society}, \bibinfo{pages}{9117--9126}.
\newblock
\urldef\tempurl%
\url{https://doi.org/10.1109/CVPR.2018.00950}
\showDOI{\tempurl}


\bibitem[Wang et~al\mbox{.}(2019)]%
        {wang2019haq}
\bibfield{author}{\bibinfo{person}{Kuan Wang}, \bibinfo{person}{Zhijian Liu}, \bibinfo{person}{Yujun Lin}, \bibinfo{person}{Ji Lin}, {and} \bibinfo{person}{Song Han}.} \bibinfo{year}{2019}\natexlab{}.
\newblock \showarticletitle{{HAQ:} {H}ardware-Aware {A}utomated {Q}uantization {W}ith {M}ixed {P}recision}. In \bibinfo{booktitle}{\emph{{IEEE} Conference on Computer Vision and Pattern Recognition, {CVPR} 2019, Long Beach, CA, USA, June 16-20, 2019}}. \bibinfo{publisher}{Computer Vision Foundation / {IEEE}}, \bibinfo{pages}{8612--8620}.
\newblock
\urldef\tempurl%
\url{https://doi.org/10.1109/CVPR.2019.00881}
\showDOI{\tempurl}


\bibitem[Wang et~al\mbox{.}(2021)]%
        {wang2021generalizable}
\bibfield{author}{\bibinfo{person}{Ziwei Wang}, \bibinfo{person}{Han Xiao}, \bibinfo{person}{Jiwen Lu}, {and} \bibinfo{person}{Jie Zhou}.} \bibinfo{year}{2021}\natexlab{}.
\newblock \showarticletitle{{G}eneralizable {M}ixed-Precision {Q}uantization via {A}ttribution {R}ank {P}reservation}. In \bibinfo{booktitle}{\emph{2021 {IEEE/CVF} International Conference on Computer Vision, {ICCV} 2021, Montreal, QC, Canada, October 10-17, 2021}}. \bibinfo{publisher}{{IEEE}}, \bibinfo{pages}{5271--5280}.
\newblock
\urldef\tempurl%
\url{https://doi.org/10.1109/ICCV48922.2021.00524}
\showDOI{\tempurl}


\bibitem[Wu et~al\mbox{.}(2018)]%
        {wu2018mixed}
\bibfield{author}{\bibinfo{person}{Bichen Wu}, \bibinfo{person}{Yanghan Wang}, \bibinfo{person}{Peizhao Zhang}, \bibinfo{person}{Yuandong Tian}, \bibinfo{person}{Peter Vajda}, {and} \bibinfo{person}{Kurt Keutzer}.} \bibinfo{year}{2018}\natexlab{}.
\newblock \showarticletitle{{M}ixed {P}recision {Q}uantization of {C}onv{N}ets via {D}ifferentiable {N}eural {A}rchitecture {S}earch}.
\newblock \bibinfo{journal}{\emph{CoRR}}  \bibinfo{volume}{abs/1812.00090} (\bibinfo{year}{2018}).
\newblock
\showeprint[arXiv]{1812.00090}
\urldef\tempurl%
\url{http://arxiv.org/abs/1812.00090}
\showURL{%
\tempurl}


\bibitem[Xu et~al\mbox{.}(2023)]%
        {xu2023generative}
\bibfield{author}{\bibinfo{person}{Shoukai Xu}, \bibinfo{person}{Shuhai Zhang}, \bibinfo{person}{Jing Liu}, \bibinfo{person}{Bohan Zhuang}, \bibinfo{person}{Yaowei Wang}, {and} \bibinfo{person}{Mingkui Tan}.} \bibinfo{year}{2023}\natexlab{}.
\newblock \showarticletitle{Generative Data Free Model Quantization with Knowledge Matching for Classification}.
\newblock \bibinfo{journal}{\emph{IEEE Transactions on Circuits and Systems for Video Technology}} (\bibinfo{year}{2023}).
\newblock


\bibitem[Yang and Jin(2021)]%
        {yang2021fracbits}
\bibfield{author}{\bibinfo{person}{Linjie Yang} {and} \bibinfo{person}{Qing Jin}.} \bibinfo{year}{2021}\natexlab{}.
\newblock \showarticletitle{{F}rac{B}its: {M}ixed {P}recision {Q}uantization via {F}ractional {B}it-Widths}. In \bibinfo{booktitle}{\emph{Thirty-Fifth {AAAI} Conference on Artificial Intelligence, {AAAI} 2021, Thirty-Third Conference on Innovative Applications of Artificial Intelligence, {IAAI} 2021, The Eleventh Symposium on Educational Advances in Artificial Intelligence, {EAAI} 2021, Virtual Event, February 2-9, 2021}}. \bibinfo{publisher}{{AAAI} Press}, \bibinfo{pages}{10612--10620}.
\newblock
\urldef\tempurl%
\url{https://ojs.aaai.org/index.php/AAAI/article/view/17269}
\showURL{%
\tempurl}


\bibitem[Yu et~al\mbox{.}(2020)]%
        {yu2020search}
\bibfield{author}{\bibinfo{person}{Haibao Yu}, \bibinfo{person}{Qi Han}, \bibinfo{person}{Jianbo Li}, \bibinfo{person}{Jianping Shi}, \bibinfo{person}{Guangliang Cheng}, {and} \bibinfo{person}{Bin Fan}.} \bibinfo{year}{2020}\natexlab{}.
\newblock \showarticletitle{{S}earch {W}hat {Y}ou {W}ant: {B}arrier {P}anelty {NAS} for {M}ixed {P}recision {Q}uantization}. In \bibinfo{booktitle}{\emph{Computer Vision - {ECCV} 2020 - 16th European Conference, Glasgow, UK, August 23-28, 2020, Proceedings, Part {IX}}} \emph{(\bibinfo{series}{Lecture Notes in Computer Science}, Vol.~\bibinfo{volume}{12354})}, \bibfield{editor}{\bibinfo{person}{Andrea Vedaldi}, \bibinfo{person}{Horst Bischof}, \bibinfo{person}{Thomas Brox}, {and} \bibinfo{person}{Jan{-}Michael Frahm}} (Eds.). \bibinfo{publisher}{Springer}, \bibinfo{pages}{1--16}.
\newblock
\urldef\tempurl%
\url{https://doi.org/10.1007/978-3-030-58545-7\_1}
\showDOI{\tempurl}


\bibitem[Yvinec et~al\mbox{.}(2023)]%
        {yvinec2023powerquant}
\bibfield{author}{\bibinfo{person}{Edouard Yvinec}, \bibinfo{person}{Arnaud Dapogny}, \bibinfo{person}{Matthieu Cord}, {and} \bibinfo{person}{Kevin Bailly}.} \bibinfo{year}{2023}\natexlab{}.
\newblock \showarticletitle{Powerquant: Automorphism search for non-uniform quantization}.
\newblock \bibinfo{journal}{\emph{arXiv preprint arXiv:2301.09858}} (\bibinfo{year}{2023}).
\newblock


\bibitem[Zhou et~al\mbox{.}(2016)]%
        {zhou2016dorefa}
\bibfield{author}{\bibinfo{person}{Shuchang Zhou}, \bibinfo{person}{Zekun Ni}, \bibinfo{person}{Xinyu Zhou}, \bibinfo{person}{He Wen}, \bibinfo{person}{Yuxin Wu}, {and} \bibinfo{person}{Yuheng Zou}.} \bibinfo{year}{2016}\natexlab{}.
\newblock \showarticletitle{{D}o{R}e{F}a-{N}et: {T}raining {L}ow {B}itwidth {C}onvolutional {N}eural {N}etworks with {L}ow {B}itwidth {G}radients}.
\newblock \bibinfo{journal}{\emph{CoRR}}  \bibinfo{volume}{abs/1606.06160} (\bibinfo{year}{2016}).
\newblock
\showeprint[arXiv]{1606.06160}
\urldef\tempurl%
\url{http://arxiv.org/abs/1606.06160}
\showURL{%
\tempurl}


\end{thebibliography}
\end{document}